\documentclass[10pt,twocolumn,letterpaper]{article}

\usepackage[pagenumbers]{style}

\usepackage{titletoc}

\usepackage{booktabs}
\usepackage{array}
\usepackage{multirow} 
\usepackage{makecell}
\usepackage{comment}

\definecolor{paperblue}{RGB}{0, 45, 114}
\usepackage[pagebackref,breaklinks,colorlinks,allcolors=paperblue]{hyperref}
\hypersetup{
  pdftitle={Longitudinal Report Generation for Multimodal Brain MRI: A Change-Focused Metric and Vision-Language System},
  pdfauthor={Krish Patel, Peirong Liu},
  pdfsubject={Computer Vision and Pattern Recognition},
  pdfkeywords={longitudinal report generation, brain MRI, vision-language models, radiology report evaluation}
}

\title{Longitudinal Report Generation for Multimodal Brain MRI:\\A Change-Focused Metric and Vision-Language System}

\author{Krish Patel $\quad$ and $\quad$ Peirong Liu\thanks{Corresponding author.} \\
~\vspace{-0.4cm}\\Department of Electrical and Computer Engineering,\\
Data Science and AI Institute,\\
Johns Hopkins University\\  
{\tt\small \{kpate156,pliu53\}@jh.edu}
}

\begin{document}
\maketitle
\begin{abstract}
Neuroradiologists rarely read a scan in isolation, since the clinical question is often what has changed. Yet existing evaluation metrics can conflate change detection with repetition of prior findings. On brain MRI, entity-overlap metrics such as RadGraph-XL F1 favor reports with incorrect interval changes over faithful paraphrases in more than 99\% of comparisons. We address this gap through change-focused evaluation and training for longitudinal reporting. We first introduce Rad-$\Delta$, a metric for interval change agreement. It is the only metric above chance in every test, reaching AUCs of 0.85 to 0.87 on scripted change errors and LLM-reader judgments where existing metrics reach at most 0.73. Beyond brain MRI, Rad-$\Delta$ leads in three of four chest X-ray tests at substantially lower cost than LLM-based metrics, and its change labels agree with radiologist-verified labels on MS-CXR-T at $\kappa = 0.89$. We also present BrainDiff, to our knowledge the first longitudinal vision-language system for brain MRI, trained with a status-first report format and a change-pretraining curriculum. On matched patient pairs, BrainDiff achieves the highest Rad-$\Delta$ macro-MCC (0.266 internal, 0.196 external), reader scores on changed pairs, and $\mathrm{F1}_{\mathrm{temp}}$ of all compared systems, retaining 98\% of internal change agreement externally. It uses 431 visual tokens per scan pair and 0.09\,s to first token, versus 3,448 and 0.43\,s for fine-tuned Qwen3-VL-8B and at least 5,172 and 2.96\,s for MedGemma-27B. Finally, in BrainDiff (brain MRI) and MAIRA-2 (chest X-ray), likelihood scoring distinguishes worsening from improvement earlier in training than generated reports do. Our training closes this gap earliest on brain MRI and improves reported change direction over standard training at every evaluated epoch in both models. Code is publicly available at \href{https://github.com/jhuldr/BrainDiff}{https://github.com/jhuldr/BrainDiff}.
\end{abstract}

\section{Introduction}
\label{sec:intro}

Radiology report generation has advanced rapidly, from CNN-RNN captioners and memory-augmented transformers~\cite{r2gen} to grounded models~\cite{rgrg} and large multimodal language models~\cite{llavamed,maira}. 
Follow-up imaging, however, requires reports that explain what has changed since a prior study. 
Longitudinal report generation has been studied extensively in chest radiography through temporal image representations~\cite{biovilt}, alignment of current and prior studies~\cite{serra,hergen}, and prior-report conditioning~\cite{maira2,mlrg}. CT2RepLong further extends this task to 3D chest CT~\cite{ct2rep}.

However, how well these systems capture interval change remains difficult to establish from the existing evaluation: follow-up reports often repeat prior findings, allowing metrics that reward shared content to \textit{obscure change errors}. This limitation is particularly relevant to brain MRI: our reports are longer than MIMIC-CXR (chest X-ray) findings (median 141 vs.\ 47 words) and contain more change language (6.2 vs.\ 1.6 change terms per report)~\cite{mimiccxr,bin_dataset,mrrate}. In our brain MRI evaluation, RadGraph-XL F1 assigns higher scores to reports with interval-change errors over meaning-preserving paraphrases in more than 99\% of comparisons. Temporal metrics also struggle: $\mathrm{F1}_{\mathrm{temp}}$ and Change-F1 distinguish scripted change errors from synonym swaps with AUCs of only 0.45 and 0.66, respectively~\cite{radgraphxl,libra,ctdeltabench}. 

The same blind spot hides a known shortcut. When a prior report is available, models may rely heavily on their text rather than compare scans, and may even hallucinate comparisons to prior studies they were never shown~\cite{ramesh}. This shortcut is rarely quantified: reported gains from ``adding the prior'' conflate genuine visual comparison with better text priors (Sec.~\ref{sec:sees}). Without an explicit control, it is difficult to say \textit{how much} a longitudinal model ``sees''.

These challenges are especially consequential for brain MRI, where longitudinal report generation remains underexplored. Recent medical imaging foundation models have advanced general-purpose visual representations~\cite{biomedclip,radfm,m3d,merlin}. In neuroimaging, NeuroVFM~\cite{neurovfm}, self-supervised on 5.24M clinical MRI/CT volumes, has demonstrated strong performance in radiologic diagnosis and single-study reporting.  Yet these capabilities do not ensure effective longitudinal comparison: our baseline trained on comparison reports alone recalls only about one quarter of worsened findings. Subtle changes within complex 3D brain anatomy and limited paired longitudinal data further complicate learning. 

We introduce Rad-$\Delta$ for interval-change agreement evaluation and BrainDiff for longitudinal brain MRI report generation. We further examine when change-sensitive information emerges during training and how generated conclusions depend on the imaging inputs. Our main contributions are:

\begin{itemize}
    \item \textbf{Rad-$\Delta$, a metric for interval-change agreement.} Rad-$\Delta$ distinguishes new, worsened, improved, stable, and resolved findings.  We evaluate Rad-$\Delta$ using scripted and LLM-written change errors, meaning-preserving edits, and LLM-reader judgments. On brain MRI, Rad-$\Delta$ detects change errors from meaning-preserving edits with AUCs of 0.85 internally and 0.87 externally, versus at most 0.73 for existing metrics, and tracks LLM-reader judgments with an AUC of 0.87. We further evaluate Rad-$\Delta$ on chest X-ray, where it leads in three of four tests and its change labels agree with radiologist-verified labels on MS-CXR-T at $\kappa = 0.89$ (Sec.~\ref{sec:metricval}). 
 
    \item \textbf{BrainDiff, to our knowledge the first longitudinal vision-language system for multimodal brain MRI.} BrainDiff combines a status-first report format with a change-pretraining curriculum to emphasize interval-change prediction. Together, these choices improve Rad-$\Delta$ agreement on both changed and stable findings and increase recall of worsened findings from 0.253 to 0.446 (Sec.~\ref{sec:ladder}). On matched patient pairs, BrainDiff obtains the highest macro-MCC and changed-pair reader scores of all compared systems and retains 98\% of its internal Rad-$\Delta$ change agreement and 74\% of its macro-MCC at an external site (Sec.~\ref{sec:comparison}). 
 
    \item \textbf{Analyses of training dynamics and image dependence.} We identify a gap between likelihood-based discrimination and generated change statements: early in training, likelihood scoring distinguishes worsening from improvement more effectively than generated reports do (AUC 0.776 vs.\ 0.647 for BrainDiff at epoch 5), and our training closes this gap earlier than the evaluated alternatives (Sec.~\ref{sec:knows}). We further use input ablations to assess sensitivity to temporal image evidence. Swapping scan order reduces BrainDiff's direction-of-change AUC from 0.714 to 0.409, a drop of 0.305 against 0.163 for the baseline without our format and curriculum (Sec.~\ref{sec:sees}). 
\end{itemize}
\section{Related Work}
\label{sec:related}

\textbf{Report generation and longitudinal modeling.} Radiology report generation has progressed from encoder--decoder and memory-augmented models~\cite{r2gen} to grounded generation~\cite{rgrg} and medical vision-language models~\cite{llavamed,maira,radvlm}. Despite these advances, hallucinated findings persist, motivating approaches that ground generated sentences in image evidence~\cite{rgrg,maira2}. Longitudinal approaches primarily target chest imaging. Chest X-ray systems learn temporal image--text representations~\cite{biovilt}, align current and prior studies before decoding~\cite{serra,hergen}, and condition on prior reports~\cite{maira2,mlrg}. Libra introduces a temporal alignment connector~\cite{libra}, while CT2RepLong incorporates the prior chest CT volume and report through cross-attention fusion~\cite{ct2rep}. Prior work also documents unsupported comparisons~\cite{ramesh} and comparable performance with generated versus real prior reports~\cite{nicolson}. These findings motivate explicit controls for visual dependence, consistent with broader concerns about shortcut learning and language priors in vision-language models~\cite{geirhos-shortcut,vqacp}. Our work complements these approaches with a status-first report format and a change-pretraining curriculum, together with input ablations that assess how generated change statements depend on the scans (Sec.~\ref{sec:sees}). 

\vspace{0.1cm} \noindent 
\textbf{3D medical foundation models.} 
With the development of 3D vision encoders, researchers have explored report generation for 3D medical images. CT2Rep introduced 3D chest-CT reporting~\cite{ct2rep}, RadFM, M3D-LaMed, Merlin, and Med3DVLM~\cite{radfm,m3d,merlin,med3dvlm} target general 3D images. Volumetric token costs are often addressed through pooling or resampling~\cite{flamingo}. Brain MRI introduces an additional challenge: multiple sequences, whose availability varies across examinations. The few recent 3D neuro foundation models, including HLIP, NeuroVFM, AutoRG-Brain, ReMIND, and BrainIAC~\cite{hlip,neurovfm,autorg,brainIAC, remind}, are built upon backbones trained and evaluated on individual studies. To our knowledge, no prior 3D report-generation system addresses interval change in brain MRI. BrainDiff uses NeuroVFM as its backbone, combining paired-study inputs with change-focused training for longitudinal reporting (Sec.~\ref{sec:architecture}).

\vspace{0.1cm} \noindent 
\textbf{Evaluation of interval change.} Report evaluation extends beyond n-gram overlap toward entity- and relation-level matching~\cite{radgraphxl}, entity-embedding similarity~\cite{ratescore}, and LLM-based clinical error assessment~\cite{green}.  Although comparison errors are recognized in report evaluation~\cite{rexval}, overall scores can mix agreement on content with agreement on change. Metrics aimed at temporal content exist for chest X-ray and CT: $\mathrm{F1}_{\mathrm{temp}}$ computes an F1 over fixed temporal keywords without explicitly linking them to the findings~\cite{libra}. 
Change-F1 uses LLM-based extraction and finding matching for chest CT~\cite{ctdeltabench}. Rad-$\Delta$ explicitly evaluates finding-level interval-change agreement. We compare it with both general report metrics and temporal metrics using controlled change errors in brain MRI, and assess its generalizability through chest X-ray evaluation (Sec.~\ref{sec:metricval}).

\begin{figure*}[t]
	\centering
	\includegraphics[width=\linewidth]{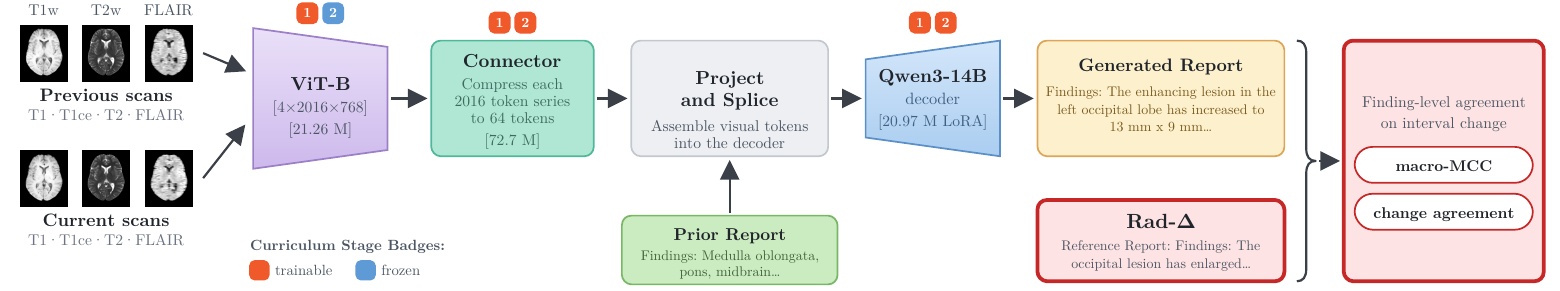}
	\caption{{\textbf{Overview of BrainDiff and Rad-$\Delta$.} Each MRI series is independently encoded by a shared ViT-B and compressed to 64 visual tokens. Stage 1 pretrains change prediction from paired scans without a prior report. Stage 2 combines prior/current visual tokens and an edited prior report in Qwen3-14B to generate a status line followed by a comparison report. Rad-$\Delta$ supplies change labels for training and evaluates finding-level interval-change agreement between generated and reference reports, excluding the generated status line.}}
	\label{fig:arch}
\end{figure*}

\section{Method}
\label{sec:method}

\subsection{Problem Setup and Overview}
\label{sec:setup}

Given two studies from one patient at times $t_1$ and $t_2$, each with up to four structural sequences (T1w, T1ce, T2w, FLAIR), and the report from $t_1$, we aim to generate a status line naming the interval changes, followed by a free-text comparison report. For each study, HD-BET skull-strips a reference series, and the remaining series are rigidly coregistered to it. All series are then mapped to the MNI-152 template using a shared affine transform in ANTs and resampled to $193 \times 229 \times 193$ voxels at 1\,mm isotropic resolution, making token locations spatially comparable across timepoints~\cite{hdbet,avants2008symmetric,fonov2011}. Series covering less than 10\% of the template or lying largely outside its bounds are excluded.

Training consists of two stages: change pretraining and report generation (Fig.~\ref{fig:arch}, Sec.~\ref{sec:curriculum}). Rad-$\Delta$ (Sec.~\ref{sec:raddelta}) provides the change labels and metrics used throughout the paper.

\subsection{Rad-$\Delta$: A Change-Focused Metric}
\label{sec:raddelta}

{Rad-$\Delta$ measures finding-level agreement on interval change between generated and reference reports.} 

\vspace{0.05cm}
\noindent \textbf{Change-aware extraction.} RadGraph-XL extracts the entities in a report and the relations between them, but it has no label for interval change. For example, “larger” is tagged as an observation and linked to “lesion” with nothing marking its change or giving it direction~\cite{radgraphxl}. Rad-$\Delta$ solves this. First, a change model assigns every Observation entity one of six statuses: \textit{NEW, WORSENED, IMPROVED, STABLE, RESOLVED, or NONE.} It is a separate encoder, initialized from RadGraph-XL and trained for this task, with a feed-forward head over span embeddings. RadGraph-XL’s own entity and relation extraction is left untouched, so its existing scores are unaffected. Second, because RadGraph-XL labels individual words, we make the finding the unit of change. In “the enhancing right frontal lesion is larger”, the words “enhancing”, “lesion” and “larger” are separate Observation entities connected by modifier relations. We group connected entities into one finding unit and assign it its most frequent change label. During training, each entity’s loss is normalized by its unit size, giving findings equal weight regardless of their length.

\vspace{0.05cm}
\noindent \textbf{Training.} Because the reports lack entity-level change annotations, we use a language model to assign silver labels to every Observation entity in 19,995 reports (10,000 brain MRI and 9,995 chest X-ray). Three fold-specific models generate out-of-fold soft targets for distillation into a single student. Distillation onto out-of-fold soft targets helps soften noise in the silver labels. On held-out reports, the student achieves unit-level macro-F1 scores of 0.902 for brain MRI and 0.906 for chest X-ray against the silver labels.

\vspace{0.05cm}
\noindent \textbf{Change agreement.} Change agreement measures whether generated and reference reports describe the same interval change for each finding. We match findings one-to-one using anatomy and word overlap via the Hungarian algorithm (Supp.~\ref{suppl:metric}). Each matched pair scores one minus the total variation distance between its change-label distributions: identical distributions score 1, confident WORSENED versus IMPROVED predictions score near 0. Reference changes omitted from the generation and generated changes absent from the reference score 0. The report score averages these contributions, with results reported separately for references with and without changed findings.

\vspace{0.05cm}
\noindent \textbf{Status presence.} For each of NEW, WORSENED, IMPROVED, RESOLVED, we mark a status as present when the noisy-OR of its unit probabilities reaches 0.5, yielding four binary labels per report. 
Macro-MCC averages Matthews correlation coefficients between generated and reference presence labels over the four statuses. We use it for model selection as constant predictions and prevalence-matched random guessing have an expected score of 0.

\subsection{BrainDiff: Longitudinal Report Generation}
\label{sec:architecture}

BrainDiff generates longitudinal reports from paired brain MRIs. Each scan is encoded and resampled independently using a vision-language backbone~\cite{neurovfm}, with visual tokens grouped into \textit{prior} and \textit{current} prompt blocks.

\vspace{0.05cm}
\noindent \textbf{Visual tokens.} Each series yields 2{,}016 tokens ($12 \times 14 \times 12$ grid, $4 \times 16 \times 16$ patches), encoded into 768-dimensional features by NeuroVFM's ViT-B~\cite{neurovfm}. We add a learned sequence embedding to each token, as the original connector confuses T1w with T1ce and T2w with FLAIR (linear-probe score: 0.69). The Perceiver resampler and projection compress each series to 64 decoder-width tokens~\cite{flamingo}, giving at most 512 visual tokens for up to eight series per pair.

\vspace{0.05cm}
{\tolerance=2000\looseness=-1 \noindent \textbf{Prior-report conditioning and decoding.} The prompt includes prior and current visual-token blocks, prior report, a statement that current report is unknown, and a generation instruction. We strip all change language from the prior report and truncate it to 384 tokens, preserving its Impression section. The Qwen3-14B decoder uses rank-16 LoRA adapters on attention projections (trainable parameters in Supp.~Tab.~\ref{tab:params})~\cite{qwen3,lora}. Test-set decoding is greedy with 1.10 repetition penalty. Validation ablations use greedy decoding without the penalty, identically across arms. Without the penalty, nearly all models repeat sentences in validation reports (Supp.~Tab.~\ref{tab:supp_loops}). With it, none exceeds 2.4\%.\par}

\subsection{Targets and the Status-First Format}
\label{sec:targets}

We develop BrainDiff using longitudinal brain MRI pairs from BIND~\cite{bin_dataset} and MR-RATE~\cite{mrrate} (details in Sec.~\ref{sec:data}).

\vspace{0.05cm}
\noindent \textbf{Targets.} For BIND, the target is the radiologist's current-study report, edited by a language model to remove non-intracranial and administrative text while preserving change statements. The LLMs are instructed to only delete words and only make minor edits for fluency. Thus, a majority of the radiologist's original language remains untouched. For MR-RATE, comparison targets are synthesized from the paired reports and restyled to match the BIND targets~\cite{mrrate}. 

Measured with Rad-$\Delta$, edited and original targets agree on the presence of any change in 95.1\% of BIND and 97.3\% of MR-RATE pairs. Separately, removing change language from BIND prior reports reduces the proportion containing change statements from 85.9\% to 2.5\%.

\vspace{0.05cm}
\noindent \textbf{Status labels.} Each longitudinal brain MRI pair receives four binary status-presence labels extracted from original current report using Rad-$\Delta$. These labels supervise the status line and serve as references for the change metrics. Comparisons with other models also include reader scores and metrics independent of Rad-$\Delta$ (Sec.~\ref{sec:expsetup}).

\vspace{0.05cm}
\noindent \textbf{Status-first format.} Each training caption begins with a status line (e.g., ``WORSENED. IMPROVED. No other interval change.'' or ``No interval change.'') followed by a line break and the report. The model thus states interval changes before describing individual findings. Training targets longer than 640 tokens are excluded. At evaluation, we remove the status line and score only the report body.

\subsection{Training Curriculum}
\label{sec:curriculum}

Training proceeds in two stages. Stage 1 trains the visual pathway and decoder to predict change statuses from paired scans alone. Stage 2 trains status-first report generation.

\vspace{0.05cm}
\noindent \textbf{Stage 1: change pretraining.} The change task predicts interval-change statuses from paired scans without a prior report, using the same status-line format as Stage 2. It uses 16{,}501 pairs (11{,}300 MR-RATE, 5{,}201 BIND) of which 53.9\% contain a change. Change and single-study tasks are sampled equally, the latter split evenly between grounding and pathology classification. The three grounding tasks share a coordinate convention and map findings to boxes, boxes to finding descriptions, region names to findings and boxes. They are built from four public lesion-segmentation datasets (ATLAS, BraTS-MEN, BraTS-METS, ISLES-22)~\cite{atlas,bratsmen,bratsmet,isles,musinguzi}. Pathology classification uses 29-label multi-label supervision on MR-RATE. We train encoder blocks 10--12, the connector, modality embeddings, and decoder LoRA for 10 epochs, selecting the checkpoint with the highest validation likelihood AUC for any change.

\vspace{0.05cm}
\noindent \textbf{Stage 2: status-first report generation.} Stage 2 initializes from Stage 1 checkpoint and minimizes cross-entropy over status line and report, with encoder frozen. Training runs for two 10-epoch cosine cycles with a warm restart~\cite{sgdr}. After each epoch, we generate reports on a fixed 256-pair validation subset. The top three checkpoints by macro-MCC are evaluated on the full validation set, and the best is retained. All ablations follow the same selection procedure.

\begin{table*}[t]
\centering
\small
\setlength{\tabcolsep}{5pt}
\caption{\textbf{Rad-$\Delta$ against existing metrics.} AUC ($\uparrow$) for separating interval-change errors from meaning-preserving controls, or reader-judged correct from incorrect reports. Cells are internal/external. Internal: held-out training-site patients; external: MIMIC-IV-Note (brain MRI) and CheXpert Plus (chest X-ray)~\cite{mimicivnote,chexpertplus}. Timing excludes model loading. Readers exclude 40 pairs GREEN could not score.}\label{tab:metric}
\vspace{-0.1cm}
\resizebox{0.9\textwidth}{!}{%
\begin{tabular}{@{}lccccccc@{}}
\toprule
Test (modality) & Rad-$\Delta$ & GREEN & rg\_er & RaTEScore & F1$_{\text{temp}}$ & Change-F1 & \makecell{Margin over best other\\(95\% CI), int / ext} \\
\midrule
Rule edits (brain MRI) & \textbf{0.85 / 0.87} & 0.65 / 0.73 & 0.64 / 0.64 & 0.61 / 0.61 & 0.45 / 0.41 & 0.66 / 0.65 & [$+$0.17, $+$0.20] / [$+$0.12, $+$0.15] \\
Written edits (brain MRI) & \textbf{0.56 / 0.66} & 0.24 / 0.15 & 0.00 / 0.00 & 0.00 / 0.01 & 0.12 / 0.14 & 0.34 / 0.37 & [$+$0.17, $+$0.27] / [$+$0.24, $+$0.34] \\
LLM readers, 960 pairs (brain MRI) & \textbf{0.87} & 0.59 & 0.46 & 0.49 & 0.55 & 0.50 & [$+$0.23, $+$0.33] \\
\midrule
Rule edits (chest X-ray) & 0.85 / \textbf{0.90} & \textbf{0.87} / 0.73 & 0.59 / 0.61 & 0.63 / 0.60 & 0.40 / 0.43 & 0.67 / 0.66 & [$-$0.03, 0.00] / [$+$0.16, $+$0.18] \\
Written edits (chest X-ray) & \textbf{0.83 / 0.65} & 0.49 / 0.43 & 0.01 / 0.00 & 0.02 / 0.04 & 0.20 / 0.12 & 0.35 / 0.29 & [$+$0.30, $+$0.38] / [$+$0.16, $+$0.26] \\
\midrule
Time per report (s) & 0.101 & 6.22 & 0.030 & 0.225 & $<$0.001 & 2.14 & -- \\
\bottomrule
\end{tabular}}
\end{table*}

\begin{figure*}[t]
	\centering
	\includegraphics[width=\linewidth]{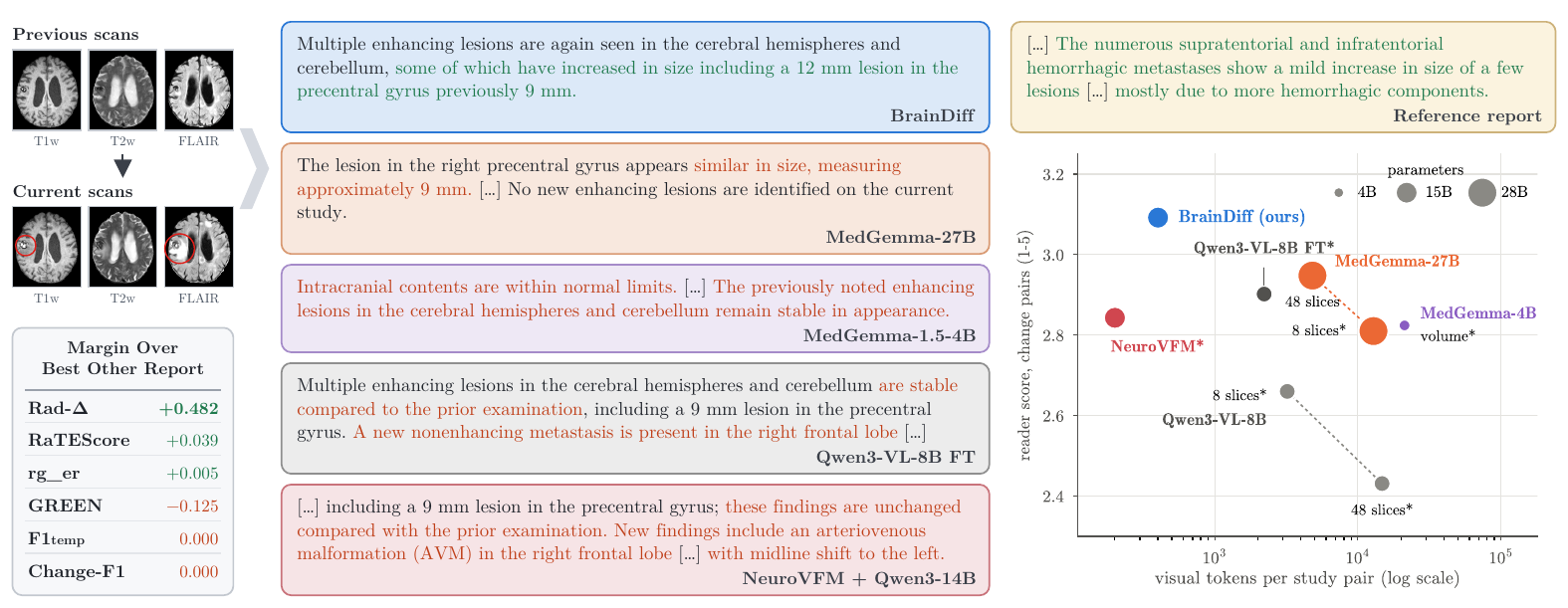}
	\caption{\textbf{Qualitative comparison and cost of change reporting.} \emph{Left and middle:} a held-out follow-up with multiple hemorrhagic metastases. Red circles mark interval hemorrhagic enlargement. The reference report records a mild increase in size of a few lesions. BrainDiff reports the interval growth (green), while baselines either describe the lesions as unchanged, overcall new disease, or hallucinate findings (red). The lower-left table values are BrainDiff minus the best of the remaining reports for each metric. \emph{Right:} change reporting against cost on the external test set. Reader score on pairs whose reference states a change (1--5) against visual tokens per study pair (log scale). Marker area is parameter count, and dashed lines join the same model given 8 and 48 slices per scan. * marks models significantly below BrainDiff. The fine-tuned Qwen3-VL-8B marker is slightly offset for visibility.}
    \label{fig:qualitative}
\end{figure*}

\subsection{What the Model Says, Knows, and Sees}
\label{sec:probes}

All three probes share 955 BIND validation pairs whose original reports indicate exactly one of WORSENED or IMPROVED under Rad-$\Delta$. Reports are generated greedily. 

\vspace{0.05cm}
\noindent \textbf{Says.} We score a generated report's stated direction as its Rad-$\Delta$ probability of WORSENED minus that of IMPROVED and report AUC on these pairs.

\vspace{0.05cm}
\noindent \textbf{Knows.} We score the direction the model's likelihoods favor. Conditioned on scans and prior report, the model scores three pairs of generic change statements, e.g., ``Interval worsening compared with the prior study'' vs. ``Interval improvement compared with the prior study''. For each study pair, we average the teacher-forced negative log-likelihood difference (improvement minus worsening) across the three statement pairs and report AUC.

\vspace{0.05cm}
\noindent \textbf{Sees.} We regenerate each report with the prior scan in both slots, the current scan in both slots, or the scans exchanged, keeping the prior report fixed. The prior-in-both and swap drops equal the original Says AUC minus the AUC under corresponding perturbation. A model relying solely on prior report would be unaffected. A drop toward chance with duplicated priors, or below chance after swapping, would be consistent with sensitivity to visual interval change.


\begin{table*}[t]
\centering
\small
\setlength{\tabcolsep}{5pt} 
\caption{\textbf{Main results.} Cells show internal/external scores; bold marks the best brain MRI result per column. All systems receive the prior report, and every generation system decodes with repetition penalty 1.10. NeuroVFM/BrainDiff have 14.9B/15.0B parameters. Fine-tuned Qwen3-VL-8B uses our pairs, with status-first targets. Other baselines are used as released. First-token latency includes vision encoding and prefill (one H200, batch 1, 50 pairs). $^{\ast}$ NeuroVFM uses two decoder passes, timed to the first token of the second. $^{\dagger}$ Not significantly different from BrainDiff (paired patient-bootstrap 95\% CI includes zero); all other baseline values differ significantly. $^{\ddagger}$ MedGemma resizes every image to 896\,px (256 tokens), so the tiles of a 48-slice montage are downsampled. Bottom rows show released chest-X-ray models scored identically on MIMIC-CXR test, for context only given the different modality and test set.}
\label{tab:main}
\resizebox{\textwidth}{!}{%
\begin{tabular}{@{}lrcccccccc@{}}
\toprule
System, visual input & \makecell{Visual tokens\\/ pair ($\downarrow$)} & \makecell{Time to first\\token (s) ($\downarrow$)} & Macro-MCC ($\uparrow$) & \makecell{Agreement,\\change ($\uparrow$)} & \makecell{Agreement,\\stable ($\uparrow$)} & \makecell{Readers: change\\pairs (1--5) ($\uparrow$)} & \makecell{Readers: stable\\pairs (1--5) ($\uparrow$)} & rg\_er ($\uparrow$) & F1$_{\text{temp}}$ ($\uparrow$) \\
\midrule
Copy prior report & -- & -- & 0.009 / 0.022 & 0.351 / 0.388 & -- & -- & -- & 0.242$^{\dagger}$ / \textbf{0.262} & 0.118 / 0.193 \\
NeuroVFM, current scan (3D) & \textbf{215} & 3.62$^{\ast}$ & 0.116 / 0.117 & 0.331 / 0.416$^{\dagger}$ & 0.798 / 0.877 & 2.11 / 2.84 & 4.20 / 4.42 & 0.213 / 0.224 & 0.321 / 0.264 \\
MedGemma-1.5-4B$^{\ddagger}$, multi-slice volume input & 20,999 & 6.01 & 0.024 / 0.029 & 0.340 / 0.345 & \textbf{0.988 / 0.990} & 2.24 / 2.82 & 4.35$^{\dagger}$ / 4.48 & 0.165 / 0.146 & 0.181 / 0.216 \\
Qwen3-VL-8B, 8 slices/scan & 3,448 & 0.38 & 0.124 / 0.090 & 0.403 / 0.429 & 0.793 / 0.744 & 2.17 / 2.66 & 3.83 / 3.77 & 0.186 / 0.184 & 0.351 / 0.279 \\
Qwen3-VL-8B, 48 slices/scan & 15,839 & 1.73 & 0.139 / 0.086 & 0.417$^{\dagger}$ / 0.438 & 0.831 / 0.723 & 2.26 / 2.43 & 4.02 / 3.65 & 0.188 / 0.185 & 0.361 / 0.280 \\
MedGemma-27B$^{\ddagger}$, 8 slices/scan & 13,792 & 15.34 & 0.077 / 0.072 & 0.395 / 0.459 & 0.829 / 0.845 & 2.35 / 2.81 & 4.05 / 4.10 & 0.215 / 0.212 & 0.328 / 0.286 \\
MedGemma-27B$^{\ddagger}$, 48 slices/scan & 5,172 & 2.96 & 0.090 / 0.081 & 0.401 / \textbf{0.464} & 0.832 / 0.865 & 2.28 / 2.95$^{\dagger}$ & 4.23$^{\dagger}$ / 4.26 & 0.227 / 0.224 & 0.312 / 0.273 \\
\midrule
Qwen3-VL-8B, fine-tuned (status-first), 8 slices/scan & 3,448 & 0.43 & 0.230 / 0.189$^{\dagger}$ & 0.372 / 0.348 & 0.939 / 0.978 & 2.57 / 2.90 & \textbf{4.45$^{\dagger}$ / 4.64} & 0.214 / 0.174 & 0.372 / 0.329 \\
\midrule
\textbf{BrainDiff (ours)}, both scans (3D) & 431 & \textbf{0.09} & \textbf{0.266 / 0.196} & \textbf{0.423} / 0.413 & 0.903 / 0.965 & \textbf{2.77 / 3.09} & 4.37 / 4.58 & \textbf{0.244} / 0.205 & \textbf{0.407 / 0.377} \\
\midrule
\multicolumn{10}{@{}l}{\textit{Chest X-ray (MIMIC-CXR test)}} \\
Libra & 1,369 & 0.22 & 0.214 & 0.395 & 0.905 & -- & -- & 0.349 & 0.362 \\
MAIRA-2 & 2,738 & 0.33 & 0.188 & 0.378 & 0.846 & -- & -- & 0.298 & 0.325 \\
\bottomrule
\end{tabular}}
\end{table*} 


\section{Experiments}
\label{sec:experiments}

\subsection{Datasets}
\label{sec:data}
Stage 2 includes 41,702 report pairs (16,627 patients), with 32,982 training pairs (18,527 from BIND~\cite{bin_dataset} I0001, 14,455 from MR-RATE)~\cite{mrrate}. BIND I0001 uses an 80/10/10 patient split, with 2,479 validation pairs (666 patients) and 2,181 internal test pairs (640 patients). BIND I0004 provides 4,060 external test pairs (1,730 patients) and is excluded from both training stages. Reported rates of any change/worsening are 40.8\%/20.6\% externally versus 54.8\%/30.3\% internally. Stage 1 uses 16,501 change pairs plus single-study grounding and pathology tasks (Sec.~\ref{sec:curriculum}).
 
\subsection{Rad-$\Delta$ Detects Change Existing Metrics Miss}
\label{sec:metricval}
Tab.~\ref{tab:metric} tests metrics on interval-change errors vs. meaning-preserving controls from same source report. Rule edits use 12 scripted error types and synonym-swap controls; written edits are LLM-generated. Reader tests use language-model judgments of generated-report correctness (Sec.~\ref{sec:expsetup}).

 \vspace{0.05cm}
\noindent\textbf{Brain MRI.} Rad-$\Delta$ is the \textit{only} metric above chance in every setting. On rule edits, it reaches 0.85 AUC internally and 0.87 externally, versus 0.66 and 0.73 for the best alternative. It further separates reports that readers judged correct from incorrect at 0.87 AUC, versus 0.59 for GREEN.

 \vspace{0.05cm}
\noindent\textbf{Written edits.} Entity-overlap metrics (rg\_er, RaTEScore, $\mathrm{F1}_{\mathrm{temp}}$) \textit{almost always favor erroneous reports} over faithful paraphrases, potentially because change errors preserve source entities while paraphrases reword them. Metrics designed for temporal content also remain below chance (F1$_{\text{temp}}$ 0.12 and 0.14, Change-F1 0.34 and 0.37). This is also Rad-$\Delta$'s weakest setting (0.56/0.66). Nevertheless, it is the \textit{most reliable} metric we tested across both modalities.

 \vspace{0.05cm}
{\tolerance=2000\looseness=-1 \noindent\textbf{Transfer.} Rad-$\Delta$ \textit{does not degrade} away from its training sites. External brain MRI AUCs match or exceed internal ones. The same checkpoint further leads three of four chest X-ray tests. Despite GREEN's lead on internal rule edits (0.87 vs.\ 0.85), it trails Rad-$\Delta$ externally (0.73 vs.\ 0.90).\par}

 \vspace{0.05cm}
\noindent\textbf{Radiologist-verified labels.} To complement the LLM-based tests, we validate change labels on MS-CXR-T~\cite{biovilt} (979 pairs, 756 patients, no patient or text overlap with change-head training data). Rad-$\Delta$'s change labels agree with the radiologist-verified, report-derived progression labels at $\kappa = 0.89$ (95\% CI 0.87--0.91), with precision at least 0.96 for every status. We map NEW to worsening and RESOLVED to improving to match its three progression labels (Supp.~\ref{suppl:mscxrt}). On brain MRI, where no such set exists, an independent labeler reproduces Rad-$\Delta$'s system rankings (Spearman 0.98 / 0.98, Sec.~\ref{sec:comparison}).

 \vspace{0.05cm}
\noindent\textbf{Cost.} Rad-$\Delta$ scores a report in 0.10\,s without LLM calls: slower than rg\_er (0.03\,s), but 21 times faster than Change-F1 (2.1\,s) and 62 times faster than GREEN (6.2\,s). Rad-$\Delta$ separates change errors better than language-model metrics in all but one test: \textit{accurate and cheap} enough to serve both for checkpoint selection in training and for final evaluation.

\subsection{Model Setup}
\label{sec:expsetup}

{\noindent\textbf{Metrics.} Change is scored against original current report using Rad-$\Delta$ macro-MCC over four statuses and change agreement, with the latter reported separately for changed and unchanged references (Sec.~\ref{sec:raddelta}). General fidelity is scored against edited target with RadGraph-XL entity and relation F1 (rg\_er) and F1$_{\text{temp}}$~\cite{radgraphxl,libra}. Comparisons use 2,000 paired patient-bootstrap resamples with 95\% intervals. All ablations use the validation data and selection procedure in Sec.~\ref{sec:curriculum}. Independent-labeler evaluation and timing protocols (single H200, batch size 1) are in Supp.~\ref{supp:eval}.

\vspace{0.05cm}
{\noindent\textbf{Compared baselines.} We compare against eight baselines on the same pairs. \textit{(i-ii)} Prior-report copying and a NeuroVFM pipeline that reports the current scan before generating a comparison with the prior report~\cite{neurovfm}. \textit{(iii-vii)} Five configurations of three released VLMs: MedGemma-1.5-4B, Qwen3-VL-8B, and MedGemma-27B~\cite{medgemma,qwen3vl, medgemma15}. The VLMs receive the same prior report and instruction. MedGemma-1.5-4B uses its recommended multi-slice volume input, and Qwen3-VL-8B and MedGemma-27B use 8 or 48 slices per scan. \textit{(viii)}~Fine-tuned Qwen3-VL-8B on the same training pairs with status-first targets (details in Supp.~\ref{supp:ft}). Test-set generation is greedy with a repetition penalty of 1.10. Detailed implementations are in Supp.~\ref{supp:baselines}.}

\vspace{0.05cm}
\noindent\textbf{Readers.} Language-model readers, from a different model family than Rad-$\Delta$’s silver labeler and blinded to system identity, rate interval-change correctness on a 1--5 scale using the prior report, the radiologist's original current report, and the candidate. Scores cover the same 500 pairs per site and are reported separately for changed and unchanged references (detailed protocols in Supp.~\ref{supp:readers}).

\subsection{Model Comparison}
\label{sec:comparison}

Tab.~\ref{tab:main} compares BrainDiff with prior-report copying, the NeuroVFM pipeline, and released and fine-tuned VLMs. BrainDiff obtains the highest macro-MCC point estimates at both sites (0.266 internal, 0.196 external), compared with 0.230 and 0.189 for the strongest baseline, the fine-tuned Qwen3-VL-8B. Released chest X-ray models reach 0.188--0.214. BrainDiff also achieves the highest language-model reader scores on changed pairs (2.77 and 3.09) and F1$_{\text{temp}}$ at both sites. Fig.~\ref{fig:qualitative} shows a representative case. The NeuroVFM pipeline reaches macro-MCCs of 0.116 and 0.117. MedGemma-1.5-4B and status-first fine-tuned Qwen score higher than BrainDiff on stable pairs, but they favor stability and score lower on changed pairs (details in Supp.~\ref{supp:comparison}).

\vspace{0.05cm}
\noindent\textbf{Matched supervision.}
With matched training pairs, targets, and labels, BrainDiff achieves a significantly higher internal macro-MCC than fine-tuned Qwen3-VL-8B (+0.036) and a numerically higher external score (0.196 vs.\ 0.189). It also achieves higher agreement on changed findings at both sites and higher external recall of worsened findings (0.264 vs.\ 0.151), with 8$\times$ fewer visual tokens.


\vspace{0.05cm}
\noindent\textbf{Slices and tokens.}
Moving from 8 single slices to 48 montaged slices per scan changes little for either MedGemma-27B (macro-MCC 0.077 to 0.090 internally and 0.072 to 0.081 externally) or Qwen3-VL-8B. The two models pay for this layout differently: Qwen3-VL encodes montages near native resolution, so its tokens rise, whereas MedGemma resizes every image to 896\,px, so its tokens fall and each slice is seen at low resolution. BrainDiff uses 12$\times$ and 32$\times$ fewer visual tokens than MedGemma-27B (431 vs.\ 5,172 and 13,792) while achieving higher macro-MCC and changed-pair reader scores at both sites (Fig.~\ref{fig:qualitative}, right).

\vspace{0.05cm}
\noindent\textbf{Readers.}
Language-model readers give BrainDiff the highest scores on changed pairs at both sites (2.77 and 3.09), corroborating its performance independently of Rad-$\Delta$. BrainDiff trails fine-tuned Qwen on unchanged pairs, so changed and stable cases should be evaluated separately.

\vspace{0.05cm}
\noindent\textbf{Independent evaluation.}
With Qwen3.6-27B as an independent status labeler, system rankings closely agree with those under Rad-$\Delta$ (Spearman 0.98 internal and external, across nine systems), and BrainDiff remains first (Supp.~Tab.~\ref{supp:labeler})~\cite{qwen36}. More details are provided in Supp.~\ref{supp:comparison}.

\vspace{0.05cm}
{\tolerance=2000\looseness=-1 \noindent\textbf{External site.}
Without site-specific adaptation, BrainDiff retains 98\% of its internal change agreement (0.413 vs.\ 0.423), while stable agreement rises from 0.903 to 0.965. Still, its macro-MCC decreases from 0.266 to 0.196.\par}

\begin{table*}[t]
\centering
\small
\setlength{\tabcolsep}{4pt}
\caption{\textbf{Training ablations: what makes the model work?} (a) Stage 1 tasks, scored by status-sentence likelihoods on validation pairs. (b) Stage 2 format and curriculum on BIND validation (2,479 pairs, 666 patients; 955 pairs for direction). One seed per setting. Markers indicate paired patient-bootstrap 95\% CIs excluding zero vs. full task set ($^{\ddagger}$) or ours ($^{\dagger}$). $^{*}$ No significant macro-MCC gain over baseline.}
\label{tab:ladder}
\vspace{-0.1cm}
\resizebox{0.9\textwidth}{!}{%
\begin{tabular}{@{}lllcccccccc@{}}
\toprule
\multicolumn{3}{@{}l}{(a) Stage 1 tasks} & \multicolumn{2}{c}{Any-change AUC} & \multicolumn{2}{c}{Worsened vs Improved} & \multicolumn{2}{c}{New vs Resolved} & \multicolumn{2}{c}{Swap drop (W vs I)} \\
\midrule
\multicolumn{3}{@{}l}{Change $+$ boxes $+$ pathology} & \multicolumn{2}{c}{0.758} & \multicolumn{2}{c}{0.747} & \multicolumn{2}{c}{0.695} & \multicolumn{2}{c}{0.380} \\
\multicolumn{3}{@{}l}{$-$ boxes} & \multicolumn{2}{c}{0.759} & \multicolumn{2}{c}{0.735} & \multicolumn{2}{c}{0.691} & \multicolumn{2}{c}{0.417$^{\ddagger}$} \\
\multicolumn{3}{@{}l}{$-$ pathology} & \multicolumn{2}{c}{0.756} & \multicolumn{2}{c}{0.692$^{\ddagger}$} & \multicolumn{2}{c}{0.641$^{\ddagger}$} & \multicolumn{2}{c}{0.294$^{\ddagger}$} \\
\multicolumn{3}{@{}l}{$-$ change task} & \multicolumn{2}{c}{0.620$^{\ddagger}$} & \multicolumn{2}{c}{0.467$^{\ddagger}$} & \multicolumn{2}{c}{0.484$^{\ddagger}$} & \multicolumn{2}{c}{$-$0.039$^{\ddagger}$} \\
\multicolumn{3}{@{}l}{$+$ trainable decoder (Stage 1-D, \textbf{ours})} & \multicolumn{2}{c}{\textbf{0.763}} & \multicolumn{2}{c}{\textbf{0.772}} & \multicolumn{2}{c}{\textbf{0.708}} & \multicolumn{2}{c}{\textbf{0.479}$^{\boldsymbol{\ddagger}}$} \\
\midrule\midrule
(b) Stage 2 & Format & Curriculum & \makecell{Macro-\\MCC} & \makecell{Agreement,\\change} & \makecell{Agreement,\\stable} & \makecell{Worsened\\recall} & \makecell{Status-line\\MCC} & \makecell{Direction\\(W-vs-I)} & \makecell{Prior-in-both\\drop} & \makecell{Swap\\drop} \\
\midrule
Baseline & Standard & None & 0.182$^{\dagger}$ & 0.408$^{\dagger}$ & 0.884$^{\dagger}$ & 0.253$^{\dagger}$ & -- & 0.633$^{\dagger}$ & 0.107$^{\dagger}$ & 0.163$^{\dagger}$ \\
$+$ format & Status-first & None & 0.197$^{\dagger}$ & 0.413$^{\dagger}$ & 0.905 & \textbf{0.456} & 0.236$^{\dagger}$ & 0.650$^{\dagger}$ & 0.083$^{\dagger}$ & 0.136$^{\dagger}$ \\
$+$ curriculum (\textbf{ours}) & Status-first & Stage 1-D & \textbf{0.236} & \textbf{0.425} & \textbf{0.915} & 0.446 & \textbf{0.269} & \textbf{0.714} & \textbf{0.197} & \textbf{0.305} \\
\midrule
Standard $+$ Stage 1-D$^{*}$ & Standard & Stage 1-D & 0.180$^{\dagger}$ & 0.413 & 0.879$^{\dagger}$ & 0.263$^{\dagger}$ & -- & 0.651$^{\dagger}$ & 0.161 & 0.231$^{\dagger}$ \\
\bottomrule
\end{tabular}}
\end{table*}

\subsection{Format and Curriculum}
\label{sec:ladder}

\noindent\textbf{Stage 1 tasks.} Tab.~\ref{tab:ladder}(a) ablates Stage 1 tasks. Removing change prediction lowers direction AUC to near chance (0.467) and the swap to $-0.039$. Removing pathology classification lowers Worsened-vs-Improved AUC from 0.747 to 0.692 and New-vs-Resolved AUC from 0.695 to 0.641. Box localization provides no significant change-detection gain but is retained for region grounding~\cite{musinguzi,maira2,rgrg}. Training the decoder yields the highest point estimates across metrics, including a swap drop of 0.479.
 
 \vspace{0.05cm}
\noindent\textbf{Format.} Tab.~\ref{tab:ladder}(b) adds the two interventions one at a time. The status-first format alone raises recall of worsened findings from 0.253 to 0.456 and stable agreement from 0.884 to 0.905, with no change in architecture or data.

 \vspace{0.05cm}
\noindent\textbf{Curriculum.} Adding Stage 1 raises macro-MCC from 0.197 to 0.236 and direction AUC from 0.650 to 0.714, with swap and prior-in-both drops of 0.305 and 0.197. The two interventions therefore appear to act on different parts of the problem. The format changes what the model says, and the curriculum affects what it takes from the images.

Stage 1 followed by standard-format training gives no significant gain over the baseline, with higher swap drop (0.231 against 0.163). Greater sensitivity to scan order thus need not translate into better generated reports, consistent with the gap between Knows and Says in Sec.~\ref{sec:knows}.

 \vspace{0.05cm}
\noindent\textbf{Transfer to chest X-ray.} To further assess generalizability, we fine-tune MAIRA-2~\cite{maira2} on MIMIC-CXR comparison pairs using a status-line warm-up followed by status-first report training (Supp.~\ref{suppl:maira}). This recipe achieves the highest point estimates across metrics and significantly improves macro-MCC over the released model (0.188 to 0.230).

\section{Analysis: Understanding Change Prediction}
\label{sec:analysis}

\begin{figure}[t]
\centering
\includegraphics[width=0.88\linewidth]{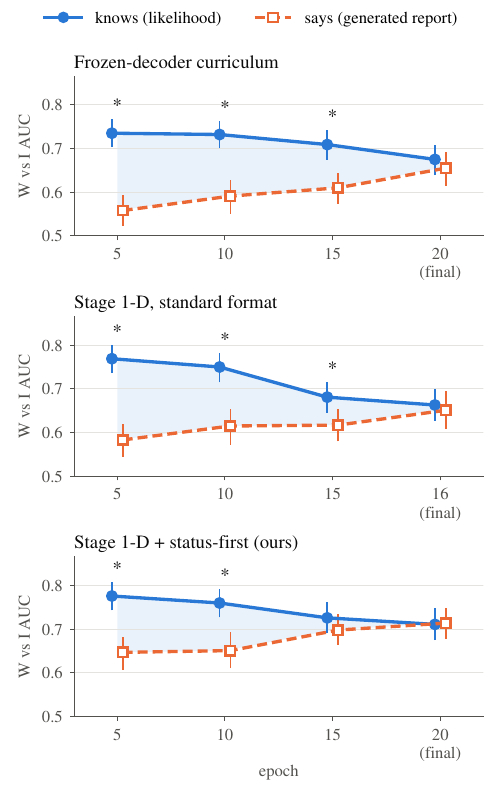}
\vspace{-0.2cm}
\caption{\textbf{Likelihood-based (Knows) and report-based (Says) direction AUC.} Worsened vs. Improved over training on 955 BIND validation pairs (95\% patient-bootstrap CIs). Shading marks gap. $*$ denotes a gap CI excluding zero. 0.5 denotes chance.}
\label{fig:knows}
\end{figure}

\begin{table}[t]
\centering
\footnotesize
\setlength{\tabcolsep}{3pt}
\caption{\textbf{Input ablations: What does the model need to see?} Models share Stage 1-D initialization, status-first training, and checkpoint selection. Only scan inputs vary (on BIND validation). 
}
\label{tab:inputs}
\resizebox{0.4\textwidth}{!}{%
\begin{tabular}{@{}lcccc@{}}
\toprule
Inputs & \makecell{\textbf{Both}\\\textbf{scans}\\\textbf{(ours)}} & \makecell{Current\\scan only\\} & \makecell{No scans\\} & \makecell{Copy\\prior\\report} \\
\midrule
macro-MCC & \textbf{0.236} & 0.226 & 0.158$^{\dagger}$ & 0.032 \\
Agree change & \textbf{0.425} & 0.408$^{\dagger}$ & 0.395$^{\dagger}$ & 0.334 \\
Agree stable & \textbf{0.915} & 0.893$^{\dagger}$ & 0.899$^{\dagger}$ & 0.994 \\
Worsened recall & 0.446 & 0.506$^{\dagger}$ & 0.419 & -- \\
Status-line MCC & \textbf{0.269} & 0.258 & 0.180$^{\dagger}$ & -- \\
W-vs-I (direction) & \textbf{0.714} & 0.646$^{\dagger}$ & n/a & -- \\
\makecell[l]{W-vs-I, prior\\replaced by current} & \makecell{0.632\\($-$0.082)} & \makecell{n/a$^{\S}$} & n/a & -- \\
Swap drop & 0.305 & 0.114$^{\ddagger}$ & n/a & -- \\
rg\_er & \textbf{0.253} & 0.251 & 0.246 & 0.246 \\
\bottomrule
\end{tabular}
}
\smallskip
\begin{minipage}{0.48\textwidth}
\scriptsize
$^{\dagger}$ Paired patient-bootstrap 95\% CI for the difference from ours excludes zero.\\
$^{\S}$ The current-only model has no prior slot, so this does not apply.\\
$^{\ddagger}$ Swapping replaces current-only input with prior scan (AUC $0.646$ to $0.532$).
\end{minipage}
\end{table}

\subsection{What the Model Sees}
\label{sec:sees}

With training and checkpoint selection fixed, the prior report alone yields macro-MCC 0.158, significantly below the full model's 0.236 but above copying (0.032). Adding the current scan reaches 0.226 (not significantly different from full), with lower direction AUC (0.646 vs.\ 0.714).

Swapping the studies reduces full model's direction AUC from 0.714 to 0.409, whereas the current-scan-only model falls from 0.646 to 0.532. The reversal in the full model supports sensitivity to the temporal ordering of both scans. 

These results suggest that paired imaging contributes particularly to \textit{direction discrimination}, while prior-report conditioning already supports substantial change prediction.

\subsection{What the Model Knows and What It Says}
\label{sec:knows}

Tracking the probes of Sec.~\ref{sec:probes} reveals that likelihood-based discrimination emerges before it is fully expressed in generated reports (Fig.~\ref{fig:knows}). At epoch 5, BrainDiff reaches 0.776 AUC on Knows but 0.647 on Says. At its selected checkpoint, scores converge to 0.711 and 0.714. This reflects both improving reports and declining likelihood discrimination.

Starting from the same Stage 1 checkpoint, status-first training achieves higher final Says AUC than standard-format training (0.714 vs.\ 0.651) and a smaller decline in Knows (0.065 vs.\ 0.106), with architecture and data fixed. The target format therefore affects both likelihood discrimination and its expression in reports. MAIRA-2 exhibits the same early gap on chest X-ray. These findings motivate tracking both probes, since generated reports alone can understate the discrimination accessible through likelihood scoring. 
Full input ablations, training trajectories, and chest-X-ray analyses are provided in Supp.~\ref{supp:analysis}.

\subsection{Limitations and Future Work}
\label{sec:limitations}

{\tolerance=2000\looseness=-1 First, Rad-$\Delta$ is trained on language-model labels and supplies both BrainDiff's supervision and evaluation labels. Human validation is limited to chest X-ray, and our readers are language models. To mitigate this, Tab.~\ref{tab:main} provides complementary checks with independent metrics and a separate labeler. A brain MRI radiologist study with independently annotated test labels would validate change labels and reader rankings while separating supervision from evaluation. Radiologist review could further assess target fidelity: BIND reports are language-model-edited, while MR-RATE targets are synthesized and used only for training.\par}

Further, BrainDiff compares two studies, and extending it to longer imaging histories would better reflect clinical follow-up. Finally, generated reports lag behind likelihood-based direction discrimination, motivating objectives that better translate this discrimination into reported conclusions.

\section{Conclusion}
\label{sec:conclusion}
\vspace{-0.07cm}
 
We presented Rad-$\Delta$, a metric for interval-change agreement, and BrainDiff, a longitudinal vision-language system for brain MRI. Rad-$\Delta$ distinguishes change errors from faithful paraphrases, while entity-overlap metrics misrank over 99\% of these comparisons. It further agrees with radiologist-verified labels on chest X-ray at $\kappa = 0.89$, and scores reports in 0.10\,s without LLM decoding, outperforming the far costlier LLM-based metrics in all but one test. Trained with a status-first report format and change-pretraining curriculum, BrainDiff achieves the highest macro-MCC point estimates, ahead of a 27B medical VLM, and retains its top rank under an independent labeler. Against a VLM fine-tuned on the same status-first targets, it achieves higher agreement on changed findings and higher external recall of worsening. Across brain MRI and chest X-ray, likelihood scoring reveals change discrimination before it is fully expressed in reports, and our training improves its expression. We believe Rad-$\Delta$ and BrainDiff advance longitudinal report generation by aligning evaluation and training with its central clinical purpose: accurately describing what has changed.  

{
    \small
    \bibliographystyle{ieeenat_fullname}
    \bibliography{main}
}

\clearpage
\appendix
\maketitlesupplementary
 
\renewcommand{\topfraction}{0.9}\renewcommand{\dbltopfraction}{0.9}\renewcommand{\bottomfraction}{0.8}
\renewcommand{\textfraction}{0.08}\renewcommand{\floatpagefraction}{0.85}\renewcommand{\dblfloatpagefraction}{0.85}
\setcounter{topnumber}{4}\setcounter{dbltopnumber}{4}\setcounter{bottomnumber}{3}\setcounter{totalnumber}{6}

\section{Model Evaluation Details}
\label{sec:supp-modelsetup}

\subsection{Evaluation and Timing}
\label{supp:eval}

\noindent\textbf{Independent labeler.}
 
To assess dependence on Rad-$\Delta$, we recompute system macro-MCC using an independent LLM labeler, Qwen3.6-27B.

\noindent\textit{Labeling prompt and decoding.} The labeler (Qwen3.6-27B, run locally with vLLM) receives the system prompt in Supp.~\ref{supp:prompt-labeler} and,
per report, a user message of the form
\begin{quote}\footnotesize\ttfamily\raggedright
PRIOR REPORT: <prior report>\\
CURRENT REPORT: <report>\\
Output:
\end{quote}
\noindent The prior report is the edited, change-free prior report that every system receives. Decoding is greedy (temperature 0, at most 48
new tokens, thinking disabled), and the output is constrained to a JSON object with four boolean keys
(\texttt{NEW}, \texttt{WORSENED}, \texttt{IMPROVED}, \texttt{RESOLVED}) by vLLM structured outputs, since only these four statuses enter Rad-$\Delta$'s macro-MCC. Each report is labeled on its own, so
the labeler never sees the reference report or the identity of the system that wrote the candidate.

The four booleans are used directly as the report's status-presence labels, the
labeler's counterpart of Rad-$\Delta$'s presence labels (Sec.~\ref{sec:raddelta}). Reference labels are the labeler's output on the original current
report. System labels are its output on each generated report (status line removed, as for Rad-$\Delta$). Macro-MCC is then computed
exactly as in Tab.~\ref{tab:main}: the Matthews correlation coefficient per status, averaged over the four statuses, with a status that a system
never or always predicts counted as 0. Outputs that do not parse would be counted as all-false; however, none occurred (0 of 62,410 labeled inputs).
Intervals use the same 2,000 patient-bootstrap resamples as Tab.~\ref{tab:main}.

\noindent\textit{Agreement with Rad-$\Delta$.} On the reference reports, the labeler agrees with Rad-$\Delta$'s presence labels at
Cohen's $\kappa$ of 0.81 / 0.88 / 0.86 / 0.81 internally and 0.78 / 0.85 / 0.87 / 0.76 externally (NEW / WORSENED / IMPROVED / RESOLVED).
Across the nine systems of Tab.~\ref{tab:main}, system macro-MCC under the two labelers differs by 0.008 on average (at most 0.022); the rank
correlation is Spearman 0.98 at both sites (Pearson 0.998 internal, 0.981 external), and BrainDiff ranks first under both labelers at
both sites.

\noindent\textbf{Statistical comparisons.}
We use paired bootstrap resampling over patients with 2,000 resamples and report 95\% confidence intervals. Resampling preserves all study pairs belonging to each sampled patient and uses the same sampled patients for both systems in each comparison. Ablations use the validation split, with every arm selected using the procedure in Sec.~\ref{sec:curriculum}.

\noindent\textbf{Timing.}
Time to first token is measured on a single NVIDIA H200 at batch size 1, with no other workload running on the GPU. Every system is evaluated on the same 50 randomly sampled internal-test pairs after three untimed warm-up pairs. We report the median latency, including visual encoding and decoder prefill. For the NeuroVFM pipeline, timing includes both decoding passes up to the first token of the comparison report.

\subsection{Trainable Parameters}
\label{supp:params}

Tab.~\ref{tab:params} lists the parameters trained in each stage.

\begin{table}[t]
\centering
\small
\caption{\textbf{Trainable parameters of BrainDiff.} Encoder blocks 1--9 and the Qwen3-14B decoder weights are frozen throughout.}
\label{tab:params}
\resizebox{0.9\linewidth}{!}{
\begin{tabular}{lrcc}
\toprule
Module & \# of Params & Stage 1-D & Stage 2 \\
\midrule
encoder blocks 10--12 & 21.26\,M & trained & frozen \\
modality embedding & 3{,}072 & trained & trained \\
connector.scan & 42.55\,M & trained & trained \\
connector.proj & 30.16\,M & trained & trained \\
decoder LoRA ($r{=}16$) & 20.97\,M & trained & trained \\
\midrule
Total trained & & 114.94\,M & 93.68\,M \\
\bottomrule
\end{tabular}
}
\end{table}

\subsection{Baseline Implementation}
\label{supp:baselines}

\noindent\textbf{Prior-report and single-study baselines.}
Copying the prior report provides a text-only reference. The NeuroVFM pipeline first uses the released single-study model to report the current scan, then uses its language model to generate a comparison against the prior report~\cite{neurovfm}.

\noindent\textbf{Released vision-language models.}
 
MedGemma-1.5-4B, MedGemma-27B, and Qwen3-VL-8B are evaluated without task-specific fine-tuning~\cite{medgemma,medgemma15,qwen3vl}. Each receives both studies as 2D images, and the same prior report, with an instruction to write the Findings section of the current report. MedGemma-1.5-4B is evaluated with the volume input recommended for MedGemma~1.5 (up to 85 axial slices per query, each as a separate image). MedGemma-27B and Qwen3-VL-8B are evaluated with 8 slices per scan, each as a separate image, and with 48-slice montages. MedGemma encodes every image at a fixed resolution with a fixed number of visual tokens, so its token count grows with the number of images. Qwen3-VL-8B's grows with image area. For MedGemma, each tile of a montage therefore receives one sixteenth of the visual tokens of a single-slice image.

\noindent\textit{Slice selection.} All inputs use only the series acquired at both timepoints (matched series), and all slices are
axial. In every format, near-empty slices (more than 90\% of pixels below $10^{-3}$ after normalization) are discarded and the
requested number of slices is taken at evenly spaced positions over the remaining ones.
\begin{itemize}
\item \textit{8 slices per scan (MedGemma-27B, Qwen3-VL-8B, fine-tuned Qwen3-VL-8B).} Each volume is min--max normalized, 8 slices are
selected, and each is resized to $224\times224$ and passed as a separate image, following MedGemma's multi-image input format. Images are ordered current study first, then prior
study, with sequences in the order T1ce, FLAIR, T2w, T1w, each preceded by a text label (e.g.\ ``Current study - FLAIR, slice 3/8''),
up to 64 images.
\item \textit{Volume input (MedGemma-1.5-4B).} Following the MedGemma~1.5 volume protocol, the volume is passed as a sequence of single slices, since the model's image encoder is 2D~\cite{medgemma15}. At most 85 slices are used per
query: $\lfloor 85 / (2 \times \text{matched series}) \rfloor$ slices per volume (42, 21, 14 or 10 for one to four matched series).
Each volume is min--max normalized, shown in radiological convention and kept at its native in-plane size (the
model's processor resizes to $896\times896$). Per volume, a label (``PRIOR examination, FLAIR: 14 axial slices, inferior to superior.'')
is followed by the slices, each followed by the text ``SLICE $j$''; prior study first, then current. A pair averages about 82 images at 256 tokens each (20,999 visual tokens) which was the most of any system.
\item \textit{48-slice montages (MedGemma-27B, Qwen3-VL-8B).} Each volume is clipped to the 0.5--99.5 percentiles of its nonzero voxels and rescaled to $[0,1]$. 48 slices are selected and resized to $224\times224$, then tiled into three $896\times896$ images of $4\times4$ slices each (slices 1--16, 17--32, 33--48, inferior to superior, read left to right and top to bottom). Images are ordered prior study first, then current, sequences in the order T1ce, FLAIR, T2w, T1w, each preceded by a label (e.g.\ ``PRIOR examination, T1w: axial slices 1-16 of 48 (whole brain, inferior to superior), read left-to-right, top-to-bottom.'').
\end{itemize}

\noindent\textit{Prompt.} Every model receives a system prompt and, in the user turn, the labeled images, then the edited prior report under the heading ``Prior report:'', and an instruction to write the Findings section for the current examination, stating for each finding whether it is new, increased/worsened, decreased/improved, resolved, or unchanged compared with the prior examination. The instruction for Qwen3-VL-8B additionally directs the model to base each comparison on the images, mirroring BrainDiff's own instruction, whereas MedGemma's is based on released prompts. All generation prompts are given in Supp.~\ref{supp:prompt-generation}.

\noindent\textit{Report sections.} Released models are instructed to write the Findings section. BrainDiff, following its training targets, also writes an Impression (87--92\% of reports), as do the fine-tuned Qwen3-VL-8B (70--84\%) and MedGemma-1.5-4B (74--82\%). Removing BrainDiff's Impression leaves its changed-pair reader scores unchanged (from 2.77 to 2.79 internally and from 3.08 to 3.07 externally; same pairs re-read together in one batch) and lowers its macro-MCC from 0.266 to 0.253 internally and from 0.196 to 0.193 externally, and its F1$_{\text{temp}}$ from 0.407 to 0.379 and from 0.377 to 0.355. All three remain above every released model at both sites, so the Impression section drives none of BrainDiff's headline results.

\noindent\textbf{Decoding.}
Test-set reports are generated greedily (temperature 0) with a repetition penalty of 1.10. The output limit is 640 new tokens for BrainDiff and 1,024 for the MedGemma and Qwen3-VL baselines, which run in vLLM with a 32,768-token context. Validation ablations use greedy decoding without a repetition penalty, identically across arms. Tab.~\ref{tab:supp_loops} reports how often outputs loop or are cut off with and without the penalty on validation.

\subsection{Fine-Tuned Qwen Baseline}
\label{supp:ft}

We fine-tune Qwen3-VL-8B using eight slices per scan and the same training pairs as BrainDiff, with status-first targets (a status line followed by the report body).

\noindent\textit{Inputs and trainable components.} Inputs are those of the Qwen3-VL-8B 8-slice row (same images, labels, order,
system prompt and instruction), with the edited prior report. Only LoRA adapters on the language model's attention projections
(query, key, value, output; rank 16, $\alpha=32$, dropout 0.05; 15.3M trainable parameters) are trained. The vision tower and
merger are frozen. The loss is computed on target tokens only.

\noindent\textit{Data.} The training pairs are BrainDiff's Stage~2 training pairs (32,982: 18,527 BIND, 14,455 MR-RATE); 285 pairs
whose target reaches 640 tokens are dropped, as for BrainDiff, leaving 32,697.

\noindent\textit{Optimization.} AdamW with learning rate $10^{-4}$, 3\% linear warm-up and cosine decay to zero, global batch 32
(4 GPUs $\times$ micro-batch 1 $\times$ 8 accumulation steps), gradient clipping at 1.0, bf16, seed 0.

\noindent\textit{Schedule and checkpoint selection.} The model first trains for 2 epochs on the status line
alone (a warm-up analogous to Stage~1), then for 3 epochs on the status line followed by the report, continuing the warm-up adapter
with a fresh optimizer and schedule. The epoch with the highest macro-MCC on 1,000 BIND validation pairs was used: epoch 1 of 3 (validation macro-MCC 0.236; untuned model 0.114). Results with confidence intervals are in Tables~\ref{tab:s_ci_int} and~\ref{tab:s_ci_ext}.

\subsection{Language-Model Reader Evaluation}
\label{supp:readers}

Language-model readers assess whether each candidate report correctly describes interval change on a 1--5 scale. We report scores separately for pairs whose reference states a change and those whose reference does not, so that high scores for asserting no change on stable pairs cannot obscure errors on changed pairs.

\noindent\textit{Readers.} Readers are Claude Opus 5.5. Each reader receives the brief in Supp.~\ref{supp:prompt-reader} and one packet of about 73 items, and returns one JSON object per item. Each item shows the edited prior report given to the systems, the radiologist's original current report as the reference, and the candidate report (BrainDiff's and the status-first model's status lines removed).

\noindent\textit{Samples.} All systems are read on the same pairs: 500 internal and 500 external test pairs, one pair per patient,
drawn at random once (seed 2) and reused for every system. Changed and unchanged pairs are defined by Rad-$\Delta$ status presence of
the original current report (any NEW, WORSENED, IMPROVED or RESOLVED): 254 / 246 internally and 153 / 347 externally. The metric comparison in Tab.~\ref{tab:metric} uses 960 of these 1,000 pairs. GREEN could not process 40, due to its token limit, and so these pairs were removed for every metric.

\noindent\textit{Blinding.} Within a batch, items of all systems in the batch are shuffled together, so readers cannot tell which
system wrote a candidate. Baselines are read in a batch together with BrainDiff.

\noindent\textit{Aggregation.} BrainDiff's score for each pair is the mean of four readings; every other system's score is its first reading. BrainDiff's single-batch means vary little (changed pairs 2.69--2.80 internally and 3.03--3.11 externally, all above every baseline; unchanged pairs 4.28--4.39 and 4.49--4.60). A random 10\% of items in each batch is read again by a different reader, only to estimate reliability (exact agreement 0.80--0.86, within one point 0.98--1.00). Paired differences against BrainDiff use patient-bootstrap intervals (2,000 resamples) over the same pairs.


\section{Model Comparison Details}
\label{supp:comparison}

\noindent\textbf{Changed and stable findings.}
 
The systems differ in their tendency to assert interval change. MedGemma-1.5-4B rarely predicts change, achieving stable agreement of 0.988 internally and 0.990 externally, but macro-MCC of only 0.024 and 0.029 and recall of worsened findings of 0.012 and 0.006. MedGemma-27B also rarely reports worsening (recall 0.03--0.06) and reaches macro-MCC 0.072--0.090. Qwen3-VL-8B asserts change more readily and has the lowest stable agreement and stable-pair reader scores. Status-first fine-tuned Qwen3-VL-8B also favors stability, with stable agreement of 0.939 and 0.978. BrainDiff leads on macro-MCC and changed-pair reader scores while ranking third among generative systems on stable agreement.

The NeuroVFM pipeline provides the closest architectural comparison, since it shares BrainDiff's encoder and decoder family but reports the current study independently before generating a textual comparison. Its macro-MCC is 0.116 internally and 0.117 externally, compared with BrainDiff's 0.266 and 0.196.

\noindent\textbf{Matched supervision.}
Compared with fine-tuned Qwen3-VL-8B, BrainDiff achieves higher change agreement internally (0.423 vs.\ 0.372) and externally (0.413 vs.\ 0.348), and higher recall of worsened findings (0.514 vs. 0.350 internally, 0.264 vs.\ 0.151 externally). Changed-pair reader scores show the same ordering: 2.77 vs.\ 2.57 internally and 3.09 vs.\ 2.90 externally. BrainDiff uses 8$\times$ fewer visual tokens. Its internal macro-MCC advantage of 0.036 is significant, whereas the external difference of 0.006 is not. Fine-tuning details are provided in Supp.~\ref{supp:ft}.

\noindent\textbf{Slice density and computational cost.}
More slices per scan does not consistently help the released models. Qwen3-VL-8B's macro-MCC changes from 0.124 to 0.139 internally and from 0.090 to 0.086 externally when moving from 8 to 48 slices, and its external reader scores decline. MedGemma-27B changes from 0.077 to 0.090 internally and from 0.072 to 0.081 externally, although its 48-slice input uses fewer visual tokens (5,172 vs.\ 13,792 per pair) because each montage is encoded as a single fixed-size image. MedGemma-1.5-4B's volume input uses the most visual tokens of any system (20,999 per pair) and reaches the lowest macro-MCC among generative systems (0.024 and 0.029). BrainDiff uses 431 visual tokens per pair and 15.0B parameters, against 27.4B for MedGemma-27B. Fig.~\ref{fig:qualitative} (right) compares external changed-pair reader scores against visual-token count and model size.

\noindent\textbf{Reader evaluation.}
Language-model readers rate BrainDiff highest on changed pairs at both sites (2.77 and 3.09). The closest competitor internally is the fine-tuned Qwen3-VL-8B (2.57); externally, it is MedGemma-27B with 48 slices (2.95). Thus, the respective margins are 0.20 and 0.14 on the five-point scale. 

On unchanged pairs, BrainDiff scores 4.37 internally and 4.58 externally, below only status-first fine-tuned Qwen3-VL-8B (4.45 and 4.64). Every system scores more than one point lower on changed pairs than on unchanged pairs, indicating that correctly describing change remains more difficult than reporting stability.

\noindent\textbf{Independent labeler.}
 
Qwen3.6-27B's status labels agree with Rad-$\Delta$ on reference reports at per-status $\kappa$ values of 0.76--0.88 (Supp.~\ref{supp:eval}). System rankings closely agree across labelers (Spearman 0.98 at both sites, over the nine systems of Tab.~\ref{tab:main}), with BrainDiff retaining the highest point estimate and the same pattern of significant margins (Supp.~Tab.~\ref{supp:labeler}). Its margin is significant against every system except status-first fine-tuned Qwen3-VL-8B at the external site ($+0.010$ [$-0.019$, $+0.041$]).

\begin{table}[!tb]
\centering\small
\caption{\textbf{Macro-MCC under an independent labeler.} Qwen3.6-27B replaces Rad-$\Delta$ as the status labeler; Rad-$\Delta$ values are repeated for comparison. Cells are internal / external test. The last column is BrainDiff minus the system under the independent labeler; all margins are significant except the one marked n.s.}
\label{supp:labeler}
\resizebox{\linewidth}{!}{%
\begin{tabular}{@{}lccc@{}}
\toprule
System & Independent labeler & Rad-$\Delta$ & BrainDiff $-$ system \\
\midrule
\textbf{BrainDiff (ours)} & \textbf{0.264 / 0.180} & 0.266 / 0.196 & -- \\
Qwen3-VL-8B, fine-tuned (status-first) & 0.219 / 0.170 & 0.230 / 0.189 & $+$0.045 / $+$0.010 (n.s.) \\
Qwen3-VL-8B, 48 slices & 0.139 / 0.100 & 0.139 / 0.086 & $+$0.125 / $+$0.080 \\
NeuroVFM $\rightarrow$ Qwen3-14B & 0.119 / 0.108 & 0.116 / 0.117 & $+$0.145 / $+$0.072 \\
Qwen3-VL-8B, 8 slices & 0.118 / 0.089 & 0.124 / 0.090 & $+$0.146 / $+$0.091 \\
MedGemma-27B, 48 slices & 0.083 / 0.082 & 0.090 / 0.081 & $+$0.181 / $+$0.098 \\
MedGemma-27B, 8 slices & 0.072 / 0.077 & 0.077 / 0.072 & $+$0.192 / $+$0.103 \\
MedGemma-1.5-4B, volume input & 0.031 / 0.033 & 0.024 / 0.029 & $+$0.233 / $+$0.148 \\
Copy prior report & 0.000 / 0.000 & 0.009 / 0.022 & $+$0.264 / $+$0.180 \\
\bottomrule
\end{tabular}}
\end{table}

\begin{table*}[!t]
\centering
\caption{\textbf{Degeneration: greedy vs.\ repetition penalty 1.1.}
1,000 BIND validation pairs (the same pairs for every system), Table~\ref{tab:main} inputs and prompts, greedy decoding
with and without the penalty; values in \%.
\emph{Loop}: a sentence of more than 30 characters occurring at least three times. \emph{Cut}: output without final punctuation
(generation reached the token limit: 640 new tokens for BrainDiff, 1,024 for the baselines). The penalty applies to generated tokens only.
For NeuroVFM $\rightarrow$ Qwen3-14B it applies to the comparison step.}
\label{tab:supp_loops}
\small
\begin{tabular}{lcccc}
\toprule
 & \multicolumn{2}{c}{Loop} & \multicolumn{2}{c}{Cut} \\
\cmidrule(lr){2-3}\cmidrule(lr){4-5}
System & Greedy & RP 1.1 & Greedy & RP 1.1 \\
\midrule
NeuroVFM $\rightarrow$ Qwen3-14B   & 0.0  & 0.0 & 0.0  & 0.0 \\
Qwen3-VL-8B, 8 slices              & 3.1  & 0.0 & 2.9  & 0.0 \\
Qwen3-VL-8B, 48 slices             & 5.0  & 0.0 & 4.7  & 0.0 \\
MedGemma-1.5-4B, volume            & 59.6 & 0.8 & 55.0 & 0.9 \\
MedGemma-27B, 8 slices             & 7.2  & 2.4 & 0.1  & 0.0 \\
MedGemma-27B, 48 slices            & 1.5  & 1.0 & 0.3  & 0.1 \\
Qwen3-VL-8B FT, status-first       & 18.0 & 0.0 & 17.6 & 0.9 \\
\midrule
BrainDiff (ours)                   & 6.0  & 0.3 & 6.0  & 1.5 \\
\bottomrule
\end{tabular}
\end{table*}

\noindent\textbf{External-site performance.}
The external site is evaluated without adaptation and contains fewer reports stating interval change. BrainDiff's change agreement remains close to its internal value (0.413 vs.\ 0.423), while stable agreement increases from 0.903 to 0.965. Macro-MCC falls from 0.266 to 0.196, showing that preserved agreement does not imply unchanged performance across all measures.

\noindent\textbf{Context from chest X-ray.}
When scored with the same metric on MIMIC-CXR, released Libra and MAIRA-2 obtain macro-MCCs of 0.214 and 0.188. These results only serve to provide context for absolute score magnitudes.


\section{Input Dependence and Training Dynamics}
\label{supp:analysis}

\subsection{Input Dependence}
\label{supp:sees}

Tab.~\ref{tab:inputs} reports input ablations trained with the same recipe and checkpoint-selection procedure. Without scans, macro-MCC reaches 0.158, compared with 0.236 for the full model and 0.032 for copying the prior report. This demonstrates substantial predictive signal from the textual prior, while the significant gap to the full model indicates an additional contribution from imaging.

\noindent\textbf{Contribution of the prior scan.}
Using the current scan and prior report yields a macro-MCC of 0.226, not significantly different from the full model's 0.236. Removing the prior scan nevertheless reduces direction AUC from 0.714 to 0.646 and lowers agreement on both changed and stable references. Recall of worsened findings rises from 0.446 to 0.506, consistent with a greater tendency to assert worsening; without the prior scan, the model can compare the current image only against the prior report's description, which may make visible lesions read as progression.

\noindent\textbf{Swapped scans.}
Swapping the full model's scans reduces direction AUC by 0.305, from 0.714 to 0.409. Under the corresponding manipulation, the current-scan-only model receives the prior image in its current slot and falls from 0.646 to 0.532. The full model's below-chance score supports sensitivity to relative temporal order, beyond the accuracy loss observed when replacing a single input image.

\noindent\textbf{Current scan in both slots.}
Replacing the prior scan with the current scan removes the visual difference between timepoints and lowers the full model's direction AUC to 0.632, close to 0.646 for the current-scan-only model. General-fidelity scores vary little across input conditions: rg\_er is 0.253 with both scans, 0.251 with only the current scan, and 0.246 without scans. These scores therefore obscure differences captured by the change-focused measures.

\noindent\textbf{Chest-X-ray comparison.}
On 534 MIMIC-CXR test pairs, swapping the released MAIRA-2 model's frontal images lowers direction AUC from 0.607 to 0.531, with a reported drop of 0.077 (95\% CI 0.023--0.130). Its output is sensitive to the imaging inputs, but the swapped score remains near chance rather than showing the reversal observed for BrainDiff.

\subsection{Likelihood and Report Discrimination}
\label{supp:knows}

\noindent\textbf{Training trajectories.}
All three models show stronger likelihood-based than report-based discrimination early in training (Fig.~\ref{fig:knows}). At epoch 5, the frozen-decoder curriculum reaches Knows AUC 0.735 and Says AUC 0.558, with a significant gap through epoch 15. By the selected checkpoint, Knows has declined to 0.675, so convergence partly reflects reduced likelihood discrimination.

BrainDiff has the highest Says AUC at epoch 5 (0.647, against 0.558 and 0.583) and the highest Knows AUC at the selected checkpoint (0.711, against 0.675 and 0.663). Its final Knows and Says scores are 0.711 and 0.714.

\noindent\textbf{Effect of target format.}
The standard-format and status-first models begin report training from the same Stage 1 checkpoint and differ only in target format. Under standard-format training, Knows declines by 0.106 and Says ends at 0.651. Under status-first training, the decline is 0.065 and Says ends at 0.714. With architecture and data fixed, these results show that the target format leads to both better preservation of likelihood discrimination and more accurate generated change statements.

\noindent\textbf{Reports with and without stated direction.}
At epoch 5, Knows exceeds Says by 0.13--0.19 AUC. Approximately half of this gap reflects reports that state no direction, for which likelihood scoring still distinguishes worsening from improvement with AUCs of 0.71--0.74. Among reports that do state a direction, Says still trails Knows by approximately 0.09, with confidence intervals excluding zero in all three models.

No model's generated reports reach its epoch-5 likelihood AUC; for BrainDiff, the corresponding scores are 0.714 and 0.776. This suggests room to improve how change-sensitive signals are expressed in reports.

\noindent\textbf{Transfer to chest X-ray.}
MAIRA-2 exhibits the same early separation between likelihood-based discrimination and generated change statements. The status-first curriculum also yields higher scores on both probes at every evaluated epoch, indicating that the observed pattern extends beyond BrainDiff and brain MRI (Supp.~\ref{suppl:maira}).

\subsection{Additional Evaluation Limitations}

Rad-$\Delta$ is validated against radiologist-verified labels on MS-CXR-T, whereas brain MRI evaluation relies on controlled edits, language-model readers, and agreement with an independent labeler. Because Rad-$\Delta$ also supplies training supervision, we additionally report reader scores, rg\_er, F1$_{\text{temp}}$, and independent-labeler results (Sec.~\ref{sec:comparison}).

These checks provide complementary evidence but do not replace a radiologist reader study, which is planned. BIND targets preserve radiologist-authored current reports after language-model editing, primarily by deletion (Supp.~\ref{suppl:rewrite}). MR-RATE comparison targets are synthesized and restricted to training. Patient-bootstrap intervals quantify variation across evaluated patients but do not capture variability across training seeds.


\section{Change Agreement: Full Definition}
\label{suppl:metric}

\paragraph{Findings and their change distributions.}
Rad-$\Delta$ extracts entities and relations with the frozen modern-radgraph-xl path and adds a change head that assigns
every Observation entity a distribution over six labels: NEW, WORSENED, IMPROVED, STABLE, RESOLVED and NONE (no interval-change
statement). A \emph{finding} is a group of Observation entities joined by \texttt{modify} links (a finding and its descriptors).
Its label is the most common label among its entities (ties go to a change label over NONE), and its distribution is the mean of
its entities' distributions. Every mention is its own finding, so a report that mentions a lesion three times has three findings.

\paragraph{Pairing.}
Each finding is described by its words (all member tokens, lower-cased) and its anatomy (Anatomy entities its members are
\texttt{located\_at}, plus Anatomy entities modifying those). The similarity of two findings is
$0.5\,J(\text{anatomy}) + 0.5\,J(\text{words})$ when both have anatomy and $J(\text{words})$ otherwise, with $J$ the Jaccard
index. Anatomy is weighted because the change word is often the finding itself (``ventricles are unchanged'' vs.\
``increased''), so the words can differ entirely while the anatomy is shared. Findings are paired one-to-one by the Hungarian
algorithm, maximizing total similarity; pairs below similarity 0.3 stay unmatched.

\paragraph{Absorbing split findings.}
Rewording often makes the extractor split or merge findings (65\% of single-word synonym swaps), and the leftover fragment would
otherwise be charged as a missed or invented change. After the one-to-one step, an unmatched finding whose words are at least
50\% contained in a finding of the other report is scored against that finding.

\paragraph{Scoring.}
STABLE and NONE probabilities are summed into one ``no change reported'' class, leaving five classes, so a pair in which one
report says ``stable'' and the other does not mention change agrees. Pairs whose two findings are both labeled NONE are dropped.
Each remaining pair scores $1 - \tfrac{1}{2}\lVert p - q \rVert_1$, one minus the total variation distance between the two
five-class distributions. Unmatched findings score 0 with weight 1 when they carry a change label (NEW, WORSENED, IMPROVED or
RESOLVED): a reference change the generation omits, or a generated change the reference does not contain. Unmatched reference
findings labeled STABLE, and unmatched findings labeled NONE, are not charged. The report score is the weighted mean of these
contributions. A report pair with nothing to score receives 1.0. Results are reported separately for references with and without
a change label.

\paragraph{How the settings were chosen.}
The pairing threshold and the absorb rule were selected on the validation split of the tagger's report pool (1,056 reports,
9,229 edits) over thresholds $\{0.3, 0.45, 0.6\}$ $\times$ absorb $\{$off, 0.5, 1.0$\}$. The selected
setting (0.3, 0.5) reached validation AUC 0.849 (absorb off: 0.829). Merging STABLE with NONE was decided on validation
paraphrases (57\% of confident label flips under rewriting were STABLE $\leftrightarrow$ NONE on the same words); it raised the
AUC for written errors vs.\ paraphrases from 0.525 to 0.678 (brain MRI) and 0.781 to 0.856 (chest X-ray) at a cost of 0.02--0.04 AUC on rule-based edits that are not interval-change errors under this definition (``The heart is
enlarged'' $\rightarrow$ ``unchanged''; deleting ``stable''). All settings were then frozen and tested once on held-out sets (Tab.~\ref{tab:metric}).

\section{Rad-$\Delta$ against Radiologist-Verified Labels (MS-CXR-T)}
\label{suppl:mscxrt}

We score the current-study report (MIMIC-CXR 2.1.0, FINDINGS and IMPRESSION) of every MS-CXR-T v1.0.0 pair with the same
Rad-$\Delta$ checkpoint that scores all BrainDiff results. MS-CXR-T progression labels are Chest ImaGenome labels extracted
automatically from the current report, each accepted or rejected by a board-certified radiologist who saw the report and the
source sentence, then matched to image pairs~\cite{biovilt}. No patients are excluded. A gold label is matched to the
tagger's findings whose tokens name the finding; if none does, it counts as \emph{not mentioned} (Tab.~\ref{tab:s_mscxrt_cov})
and is excluded from agreement. A finding's predicted status is its most frequent non-NONE label (ties to the larger margin).
\emph{Strict}: exact match. \emph{Lenient}: NEW counts as worsening and RESOLVED as improving, since
MS-CXR-T has three classes and does not document how new or resolved findings were labeled. All 135 findings the tagger labels NEW are gold worsening. All intervals: patient bootstrap, 2,000 resamples.

\begin{table*}[!t]
\centering\small
\caption{\textbf{MS-CXR-T: overlap with the tagger's training data.}}
\label{tab:s_mscxrt_overlap}
\resizebox{0.95\textwidth}{!}{%
\begin{tabular}{ll}
\toprule
Tagger training pools & 10,008 brain MRI + 10,000 chest X-ray reports (BIND) \\
Patient overlap & none possible (MIMIC-CXR: different hospital and patient-ID space) \\
MS-CXR-T reports checked & 970 current-study reports \\
Exact duplicates of a training report & 0 \\
Shared word 8-grams per report, chest X-ray pool (median / p90 / p99 / max) & 0.000 / 0.029 / 0.090 / 0.150 \\
Same, both pools & 0.000 / 0.030 / 0.106 / 0.150 \\
Reports sharing $\geq$30\% of their 8-grams & 0 \\
Reports sharing any 8-gram & 236 (24\%); median shared fraction 0.026 \\
\bottomrule
\end{tabular}
}
\end{table*}

\begin{table*}[!t]
\centering\small
\caption{\textbf{MS-CXR-T: coverage.}}
\label{tab:s_mscxrt_cov}
\resizebox{0.95\textwidth}{!}{%
\begin{tabular}{ll}
\toprule
Pairs / patients & 1,045 / 800 \\
Pairs whose current report has FINDINGS or IMPRESSION & 979 (970 current studies), 756 patients \\
Gold labels & 1,237: worsening 515, stable 490, improving 232 \\
Label quality & one expert 882, multiple experts 198, expert disagreement 157 \\
Finding named in the report & 1,146 / 1,237 (92.6\%): improving 95.7\%, stable 93.5\%, worsening 90.5\% \\
Not testable & NEW vs.\ WORSENED, RESOLVED vs.\ IMPROVED, NONE (no gold class) \\
\bottomrule
\end{tabular}
}
\end{table*}

\begin{table*}[!t]
\centering\small
\caption{\textbf{MS-CXR-T: agreement} (1,146 mentioned labels; 95\% CI).}
\label{tab:s_mscxrt_agree}
\resizebox{0.85\textwidth}{!}{%
\begin{tabular}{lll}
\toprule
 & Agreement & Cohen's $\kappa$ \\
\midrule
Strict (6 statuses vs.\ 3 gold) & 0.810 [0.785, 0.833] & 0.732 [0.701, 0.762] \\
Lenient & 0.928 [0.913, 0.944] & 0.892 [0.869, 0.914] \\
Change vs.\ no change & recall 0.916 [0.894, 0.936], precision 0.984 [0.972, 0.994] & 0.878 [0.849, 0.908] \\
\bottomrule
\end{tabular}
}
\end{table*}

\begin{table*}[!t]
\centering\small
\caption{\textbf{MS-CXR-T: per gold status.} Strict worsening agreement is low because the tagger labels 135 of the 466
worsening findings NEW (Table~\ref{tab:s_mscxrt_conf}).}
\label{tab:s_mscxrt_status}
\begin{tabular}{lrlll}
\toprule
Gold status & $n$ & Strict agreement & Lenient agreement & Lenient precision ($n$ predicted) \\
\midrule
Improving & 222 & 0.968 [0.943, 0.990] & 0.973 [0.950, 0.991] & 0.960 [0.927, 0.986] (225) \\
Stable & 458 & 0.954 [0.931, 0.974] & 0.954 [0.931, 0.974] & 0.998 [0.993, 1.000] (438) \\
Worsening & 466 & 0.592 [0.546, 0.639] & 0.882 [0.852, 0.912] & 0.990 [0.980, 0.998] (415) \\
\bottomrule
\end{tabular}
\end{table*}

\begin{table*}[!t]
\centering\small
\caption{\textbf{MS-CXR-T: confusion matrix}, all 1,237 labels (rows: gold; columns: Rad-$\Delta$).}
\label{tab:s_mscxrt_conf}
\begin{tabular}{lrrrrrrr}
\toprule
Gold & IMPROVED & RESOLVED & STABLE & WORSENED & NEW & NONE & not mentioned \\
\midrule
Improving (232) & 215 & 1 & 0 & 1 & 0 & 5 & 10 \\
Stable (490) & 7 & 0 & 437 & 3 & 0 & 11 & 32 \\
Worsening (515) & 2 & 0 & 1 & 276 & 135 & 52 & 49 \\
\bottomrule
\end{tabular}
\end{table*}

\begin{table}[!tb]
\centering\small
\caption{\textbf{MS-CXR-T: per finding and per label quality} (lenient).}
\label{tab:s_mscxrt_finding}
\resizebox{0.45\textwidth}{!}{%
\begin{tabular}{lrll}
\toprule
 & $n$ & Agreement & $\kappa$ \\
\midrule
Pneumothorax & 190 & 0.989 [0.972, 1.000] & 0.982 [0.953, 1.000] \\
Pleural effusion & 378 & 0.955 [0.933, 0.974] & 0.929 [0.896, 0.959] \\
Edema & 240 & 0.946 [0.916, 0.972] & 0.919 [0.876, 0.957] \\
Consolidation & 168 & 0.923 [0.880, 0.959] & 0.878 [0.811, 0.934] \\
Pneumonia & 170 & 0.782 [0.720, 0.843] & 0.646 [0.553, 0.738] \\
\midrule
One expert & 803 & 0.910 [0.889, 0.931] & 0.866 [0.835, 0.896] \\
Multiple experts & 191 & 0.958 [0.928, 0.984] & 0.934 [0.886, 0.975] \\
Expert disagreement & 152 & 0.987 [0.967, 1.000] & 0.979 [0.945, 1.000] \\
\bottomrule
\end{tabular}
}
\end{table}

\paragraph{Scope.}
On a second modality with no patient or text overlap with its training data, the tagger's change labels agree with
radiologist-verified labels at $\kappa = 0.89$ (lenient); its change calls are rarely wrong (precision $\geq 0.96$ per status),
and most errors are missed worsening. The gold labels are report-derived, so this measures agreement on what the report states,
not on image findings. It does not validate the brain labels directly, nor test NEW vs.\ WORSENED, RESOLVED vs.\ IMPROVED or NONE.
For 7.4\% of gold labels, no finding was matched in FINDINGS and IMPRESSION; the source sentence may sit in another section, or
the finding keywords may miss it.

\section{Main Results with 95\% Confidence \mbox{Intervals}}
\label{suppl:ci}

Tab.~\ref{tab:s_ci_int} and Tab.~\ref{tab:s_ci_ext} repeat Tab.~\ref{tab:main} with 95\% intervals: percentile patient bootstrap, 2,000 resamples, per split (internal 2,181 pairs / 640 patients; external 4,060 pairs / 1,730 patients). Reader columns use 500 pairs per site. Baselines are read together with BrainDiff. 

\begin{table*}[!t]
\centering\scriptsize
\setlength{\tabcolsep}{3pt}
\caption{\textbf{Internal test (BIND), 95\% CI.} Readers: 254 change / 246 stable references.}
\label{tab:s_ci_int}
\resizebox{\textwidth}{!}{%
\begin{tabular}{lccccccc}
\toprule
System & Macro-MCC & Agree, change & Agree, stable & Readers, change & Readers, stable & rg\_er & F1temp \\
\midrule
Copy prior report & 0.009 [$-$0.010, 0.029] & 0.351 [0.332, 0.370] & 0.988 [0.982, 0.992] & -- & -- & 0.242 [0.233, 0.250] & 0.118 [0.105, 0.132] \\
NeuroVFM $\rightarrow$ Qwen3-14B & 0.116 [0.087, 0.146] & 0.331 [0.311, 0.351] & 0.798 [0.770, 0.824] & 2.11 [1.97, 2.26] & 4.20 [4.04, 4.36] & 0.213 [0.205, 0.220] & 0.321 [0.308, 0.334] \\
Qwen3-VL-8B, 8 sl. & 0.124 [0.099, 0.150] & 0.403 [0.386, 0.420] & 0.793 [0.767, 0.818] & 2.17 [2.02, 2.33] & 3.83 [3.63, 4.03] & 0.186 [0.179, 0.193] & 0.351 [0.337, 0.365] \\
Qwen3-VL-8B, 48 sl. & 0.139 [0.111, 0.166] & 0.417 [0.399, 0.435] & 0.831 [0.805, 0.855] & 2.26 [2.11, 2.41] & 4.02 [3.83, 4.19] & 0.188 [0.181, 0.196] & 0.361 [0.346, 0.374] \\
MedGemma-1.5-4B, volume & 0.024 [0.001, 0.045] & 0.340 [0.321, 0.358] & 0.988 [0.982, 0.993] & 2.24 [2.11, 2.38] & 4.35 [4.21, 4.48] & 0.165 [0.157, 0.173] & 0.181 [0.170, 0.194] \\
MedGemma-27B, 8 sl. & 0.077 [0.052, 0.100] & 0.395 [0.377, 0.412] & 0.829 [0.807, 0.849] & 2.35 [2.21, 2.50] & 4.05 [3.88, 4.20] & 0.215 [0.207, 0.221] & 0.328 [0.313, 0.342] \\
MedGemma-27B, 48 sl. & 0.090 [0.066, 0.113] & 0.401 [0.384, 0.418] & 0.832 [0.812, 0.851] & 2.28 [2.13, 2.42] & 4.23 [4.07, 4.37] & 0.227 [0.219, 0.235] & 0.312 [0.298, 0.325] \\
Qwen3-VL-8B FT, status-first & 0.230 [0.197, 0.260] & 0.372 [0.354, 0.391] & 0.939 [0.923, 0.953] & 2.57 [2.43, 2.72] & 4.45 [4.31, 4.59] & 0.214 [0.205, 0.223] & 0.372 [0.354, 0.388] \\
\midrule
BrainDiff (ours) & 0.266 [0.234, 0.297] & 0.423 [0.405, 0.441] & 0.903 [0.883, 0.921] & 2.77 [2.62, 2.92] & 4.37 [4.24, 4.50] & 0.244 [0.235, 0.252] & 0.407 [0.390, 0.424] \\
\bottomrule
\end{tabular}
}
\end{table*}

\begin{table*}[!t]
\centering\scriptsize
\setlength{\tabcolsep}{3pt}
\caption{\textbf{External test, 95\% CI.} Readers: 153 change / 347 stable references.}
\label{tab:s_ci_ext}
\resizebox{\textwidth}{!}{%
\begin{tabular}{lccccccc}
\toprule
System & Macro-MCC & Agree, change & Agree, stable & Readers, change & Readers, stable & rg\_er & F1temp \\
\midrule
Copy prior report & 0.022 [0.003, 0.042] & 0.388 [0.369, 0.407] & 0.991 [0.988, 0.993] & -- & -- & 0.262 [0.255, 0.269] & 0.193 [0.179, 0.207] \\
NeuroVFM $\rightarrow$ Qwen3-14B & 0.117 [0.093, 0.142] & 0.416 [0.400, 0.433] & 0.877 [0.864, 0.889] & 2.84 [2.63, 3.06] & 4.42 [4.29, 4.54] & 0.224 [0.219, 0.230] & 0.264 [0.254, 0.275] \\
Qwen3-VL-8B, 8 sl. & 0.090 [0.072, 0.110] & 0.429 [0.414, 0.445] & 0.744 [0.728, 0.760] & 2.66 [2.44, 2.88] & 3.77 [3.60, 3.94] & 0.184 [0.179, 0.189] & 0.279 [0.267, 0.290] \\
Qwen3-VL-8B, 48 sl. & 0.086 [0.067, 0.107] & 0.438 [0.422, 0.454] & 0.723 [0.706, 0.740] & 2.43 [2.22, 2.64] & 3.65 [3.47, 3.82] & 0.185 [0.179, 0.190] & 0.280 [0.268, 0.291] \\
MedGemma-1.5-4B, volume & 0.029 [0.008, 0.051] & 0.345 [0.328, 0.362] & 0.990 [0.987, 0.993] & 2.82 [2.63, 3.03] & 4.48 [4.38, 4.58] & 0.146 [0.141, 0.152] & 0.216 [0.204, 0.229] \\
MedGemma-27B, 8 sl. & 0.072 [0.055, 0.089] & 0.459 [0.443, 0.475] & 0.845 [0.832, 0.858] & 2.81 [2.60, 3.01] & 4.10 [3.96, 4.23] & 0.212 [0.206, 0.219] & 0.286 [0.275, 0.297] \\
MedGemma-27B, 48 sl. & 0.081 [0.062, 0.101] & 0.464 [0.448, 0.480] & 0.865 [0.854, 0.877] & 2.95 [2.74, 3.17] & 4.26 [4.13, 4.38] & 0.224 [0.218, 0.231] & 0.273 [0.263, 0.284] \\
Qwen3-VL-8B FT, status-first & 0.189 [0.161, 0.216] & 0.348 [0.330, 0.366] & 0.978 [0.973, 0.982] & 2.90 [2.69, 3.12] & 4.64 [4.54, 4.74] & 0.174 [0.168, 0.180] & 0.329 [0.316, 0.341] \\
\midrule
BrainDiff (ours) & 0.196 [0.169, 0.221] & 0.413 [0.397, 0.429] & 0.965 [0.958, 0.972] & 3.09 [2.88, 3.30] & 4.58 [4.48, 4.67] & 0.205 [0.200, 0.211] & 0.377 [0.363, 0.391] \\
\bottomrule
\end{tabular}
}
\end{table*}

\section{Latency}
\label{suppl:latency}

\paragraph{Protocol.}
50 random BIND internal-test pairs (seed 0) after 3 untimed warm-up pairs, batch 1, one NVIDIA H200, medians. Decoding as in
Tab.~\ref{tab:main} (greedy, repetition penalty 1.1 on generated tokens). \emph{Time to first token} (TTFT) covers the vision encoder(s), the
connector or projector and language-model prefill up to the first output token; disk reads and fixed preprocessing (NIfTI to 3D
tokens, NIfTI to PNG slices) are excluded for every system. Our model runs in Hugging Face \texttt{generate} (TTFT = generation of
one token); the VLM baselines run in vLLM 0.30 with streaming and prefix caching off, one server at a time. NeuroVFM
$\rightarrow$ Qwen3-14B is two-stage, since the comparison report cannot start before NeuroVFM's own current-scan report is complete. Thus,
its TTFT is the full step-1 generation (released pipeline and decoding: beam 4, repetition penalty 1.2, Hugging Face) plus the
step-2 TTFT (vLLM). The chest X-ray models are timed with the same protocol on 50 MIMIC-CXR test pairs in Hugging Face, and image reads are excluded. TTFT is the like-for-like measure of input cost, and running ours in Hugging Face, if anything, disadvantages it.

\begin{table}[!tb]
\centering\small
\caption{\textbf{First-token latency per study pair} (median of 50 pairs, batch 1, one H200).}
\label{tab:s_latency}
\resizebox{\linewidth}{!}{%
\begin{tabular}{llc}
\toprule
System & Engine & TTFT (s) \\
\midrule
\multicolumn{3}{l}{\emph{Brain MRI (BIND internal test)}} \\
NeuroVFM $\rightarrow$ Qwen3-14B & HF + vLLM & 3.62$^{a}$ \\
Qwen3-VL-8B, 8 slices & vLLM & 0.380 \\
Qwen3-VL-8B, 48 slices & vLLM & 1.73 \\
Qwen3-VL-8B FT, status-first & vLLM + LoRA & 0.429 \\
MedGemma-1.5-4B, volume & vLLM & 6.01 \\
MedGemma-27B, 8 slices & vLLM & 15.3 \\
MedGemma-27B, 48 slices & vLLM & 2.96 \\
BrainDiff (ours) & HF & \textbf{0.089} \\
\midrule
\multicolumn{3}{l}{\emph{Chest X-ray (MIMIC-CXR test), scale context only}} \\
MAIRA-2 & HF & 0.326 \\
Libra & HF & 0.216 \\
\bottomrule
\end{tabular}
}
\\[2pt]{\footnotesize $^{a}$ Step 1 (NeuroVFM's own current-scan report) 3.58 s + step 2 TTFT 0.043 s.}
\end{table}

MAIRA-2 uses its native inputs (current lateral view when present, 24 of 50 pairs);
Libra uses its official prompt with the current and prior frontal views.

\section{Curriculum and Status-First on a Chest \mbox{X-Ray} Model (MAIRA-2)}
\label{suppl:maira}

\paragraph{Setup.}
MAIRA-2 is fine-tuned on MIMIC-CXR comparison pairs. M0: released MAIRA-2. All arms receive the current and prior frontal images only, matching BrainDiff's input structure of one current and one prior study. With its native inputs (current lateral view when available, 816 / 1,827 pairs), released MAIRA-2 scores macro-MCC 0.204 (difference from M0 $+$0.016 [$-$0.006, $+$0.039], n.s.); latency (Tab.~\ref{tab:s_latency}) is timed with native inputs. M1: plain fine-tuning (standard report target, 3 epochs).
M2: curriculum + status-first, a 2-epoch warm-up on the status line alone (Stage~1 analogue) followed by 3 epochs on status
line + report. Both fine-tuned arms keep RAD-DINO frozen, train the projector fully (lr $2\times10^{-5}$) and the language model
with LoRA rank 32 (lr $10^{-4}$), batch 30, seed 42. The epoch is chosen by validation macro-MCC (707 pairs; M1 epoch 2, M2
epoch 2) and the test set is scored once, with M2's status line removed. Test: 1,827 MIMIC-CXR pairs / 258 patients.
Differences: paired patient bootstrap, 2,000 resamples. M2$-$M1 is the combined effect of the curriculum and the format.

\begin{table*}[!t]
\centering\scriptsize
\setlength{\tabcolsep}{3pt}
\caption{\textbf{MAIRA-2, MIMIC-CXR test.} Arm rows: bold = best. Difference rows: bold = 95\% CI excludes 0.}
\label{tab:s_maira}
\resizebox{\textwidth}{!}{%
\begin{tabular}{lcccccc}
\toprule
Arm & Macro-MCC & Agree, change & Agree, stable & Worsened recall & rg\_er & F1temp \\
\midrule
M0 released MAIRA-2 & 0.188 & 0.378 & 0.846 & 0.247 & 0.298 & 0.325 \\
M1 plain fine-tuning & 0.215 & 0.406 & 0.854 & 0.241 & 0.340 & 0.363 \\
M2 curriculum + status-first & \textbf{0.230} & \textbf{0.408} & \textbf{0.870} & \textbf{0.263} & \textbf{0.342} & \textbf{0.374} \\
\midrule
M2 $-$ M1 & $+0.015$ [$-0.014$, $+0.043$] & $+0.003$ [$-0.012$, $+0.017$] & $+0.016$ [$-0.002$, $+0.033$] & $+0.022$ [$-0.027$, $+0.068$] & $+0.003$ [$-0.003$, $+0.009$] & $+0.010$ [$-0.004$, $+0.024$] \\
M2 $-$ M0 & \textbf{+0.042 [+0.010, +0.072]} & \textbf{+0.031 [+0.013, +0.048]} & \textbf{+0.024 [+0.002, +0.046]} & $+0.016$ [$-0.031$, $+0.058$] & \textbf{+0.044 [+0.037, +0.052]} & \textbf{+0.049 [+0.032, +0.066]} \\
M1 $-$ M0 & $+0.027$ [$-0.006$, $+0.056$] & \textbf{+0.028 [+0.009, +0.045]} & $+0.008$ [$-0.011$, $+0.025$] & $-0.006$ [$-0.060$, $+0.044$] & \textbf{+0.042 [+0.035, +0.049]} & \textbf{+0.038 [+0.020, +0.056]} \\
\bottomrule
\end{tabular}
}
\end{table*}

M2's status line is well formed in 100\% of reports (status-line macro-MCC 0.260), and its report agrees with its own status line
in 78\%.

\begin{table}[!tb]
\centering\small
\caption{\textbf{Knows vs.\ says on MAIRA-2.} Worsened-vs-Improved AUC on pairs whose reference states exactly one of the two.
Knows: likelihood of short content-free worsening vs.\ improvement statements (as Fig.~\ref{fig:knows}); says: Rad-$\Delta$
P(Worsened) $-$ P(Improved) of the generated report (greedy). Gap = knows $-$ says; bold gap = 95\% CI excludes 0. 168 validation
pairs (91 Worsened / 77 Improved).}
\label{tab:s_maira_ks}
\resizebox{\linewidth}{!}{%
\begin{tabular}{llccc}
\toprule
Arm & Epoch & Knows & Says & Gap \\
\midrule
M1 plain fine-tuning & 1 & 0.672 & 0.559 & \textbf{+0.113 [+0.012, +0.210]} \\
 & 2 (chosen) & 0.650 & 0.561 & $+0.088$ [$-0.021$, $+0.206$] \\
 & 3 & 0.618 & 0.555 & $+0.064$ [$-0.054$, $+0.188$] \\
\midrule
M2 curriculum + status-first & 1 & 0.692 & 0.577 & \textbf{+0.115 [+0.005, +0.223]} \\
 & 2 (chosen) & 0.682 & 0.627 & $+0.055$ [$-0.052$, $+0.165$] \\
 & 3 & 0.660 & 0.594 & $+0.067$ [$-0.026$, $+0.164$] \\
\bottomrule
\end{tabular}
}
\end{table}

\begin{figure}[!tb]
\centering
\includegraphics[width=\linewidth]{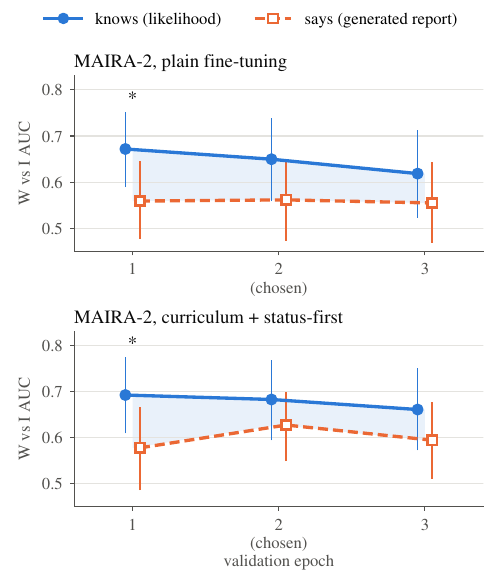}
\caption{\textbf{Knows vs.\ says on MAIRA-2}, in the layout of Fig.~\ref{fig:knows}. Knows: AUC of the likelihood the model assigns to
``worsening'' vs.\ ``improvement'' statements; says: AUC of the direction stated in its generated report (greedy). Worsened vs.\
Improved on 168 MIMIC-CXR validation pairs per epoch; 95\%
patient-bootstrap CIs; shading = the gap on validation; $\ast$ = gap CI excludes 0; dotted line = chance. As on brain MRI, both
models know the direction better than they say it at the first epoch; curriculum + status-first is higher on both throughout.}
\label{fig:s_maira_ks}
\end{figure}

\paragraph{Reading.}
M2 is best on every metric, but no single difference against plain fine-tuning is significant (macro-MCC $+0.015$ [$-0.014$,
$+0.043$]); only M2 significantly beats the released model on macro-MCC and stable agreement. As on brain MRI, both arms know the
direction better than they say it early in fine-tuning (significant at epoch 1), and the gap narrows by the chosen epoch.

\section{Silver Labels and Distillation for Rad-$\Delta$}
\label{suppl:silver}

Reports carry no entity-level change annotations, so a language model labels them. \texttt{modern-radgraph-xl} first extracts entities from each report. Every Observation entity is marked inline with a numeric id, and the language model (\texttt{gpt-6-sol}, low reasoning effort) assigns one of six attributes to each id: NEW, WORSENED, IMPROVED, STABLE, RESOLVED or NONE. It never chooses entity boundaries. The labeler sees the current report alone, from the Findings section onward, and is instructed to label only change the report states (prompt in Supp.~\ref{supp:prompt-silver}). There is no output schema; a report is kept only if every id appears exactly once with a valid attribute.

The brain MRI pool has 10,008 BIND reports and the chest X-ray pool 10,000, all from training and validation patients. In each pool 80\% are follow-up exams and 20\% are first exams, with at most two reports per patient (Tab.~\ref{tab:s_silver_data}). Training patients are split into three folds. Each fold model is trained on two folds with the silver labels and predicts the third, giving every training entity a probability distribution over the six attributes from a model that never saw it. The student is trained only on these distributions (KL divergence, entities weighted by 1/unit size). On validation reports, which also select the epoch, it reaches unit-level macro-F1 of 0.902 (brain MRI) and 0.906 (chest X-ray) against the silver labels.

Folds are assigned per patient, with brain MRI and chest X-ray reports pooled. In the student, the \texttt{modern-radgraph-xl} encoder and its entity and relation heads are frozen, so the extracted graph is identical to RadGraph-XL. The change head is a 2-layer MLP (hidden size 150, dropout 0.4) on a trainable copy of the encoder (149.0M parameters). Training uses AdamW with learning rates of $5\times10^{-5}$ for the encoder copy and $10^{-3}$ for the head, 10\% linear warm-up followed by linear decay, and up to 20 epochs with patience 5; epoch 15 was selected.

\begin{table*}[!t]
\centering\small
\caption{\textbf{Silver-label data for the Rad-$\Delta$ tagger.} Reports come from BIND sites I0001 (brain MRI) and I0007 (chest X-ray); test-split patients (Rad-$\Delta$'s patient split) are excluded before sampling.
Follow-up: the patient has an earlier exam of the same modality. At most two reports per patient. Splits are by patient (80/10/10 hash). Student macro-F1 is unit-level agreement with the silver labels on the validation
reports, which also select the epoch.}
\label{tab:s_silver_data}
\begin{tabular}{lcc}
\toprule
 & Brain MRI & Chest X-ray \\
\midrule
Sampled reports (follow-up / first exam) & 10,008 (8,006 / 2,002) & 10,000 (8,000 / 2,000) \\
With $\geq$1 Observation entity (sent to the labeler) & 10,000 & 9,995 \\
Passed the output check (used) & 9,999 & 9,994 \\
Train / validation reports & 8,943 / 1,056 & 8,873 / 1,121 \\
Train / validation patients & 6,794 / 795 & 8,288 / 1,044 \\
Train reports per fold (1 / 2 / 3) & 3,009 / 2,891 / 3,043 & 2,989 / 2,921 / 2,963 \\
Labeled Observation entities & 351,676 & 96,816 \\
\quad NEW / WORSENED / IMPROVED (\%) & 3.1 / 6.1 / 3.5 & 2.5 / 2.9 / 2.7 \\
\quad STABLE / RESOLVED / NONE (\%) & 28.7 / 0.5 / 58.1 & 21.2 / 0.7 / 70.0 \\
Student unit macro-F1 vs.\ silver (validation) & 0.902 & 0.906 \\
\bottomrule
\end{tabular}
\end{table*}

\section{Reader Study: Overall Quality}
\label{suppl:readers}

\paragraph{Setup.}
50 BIND validation pairs (one per patient, seed 1). Two systems: BrainDiff (status-first, epoch 20, greedy; status line removed)
and a standard-format model (epoch 20). The 100 reports were shuffled and
blinded into four packets of 25; each packet was read by two independent LLM readers (Claude Opus 5.5, eight readers in total). Each
item shows the edited prior report (the model's input), the radiologist's original current report as the reference, and the
candidate; only intracranial findings count, and wording and style are ignored. Readers rated overall quality (1--5) and counted, per report, invented findings (findings the reference
does not support), incorrect stable claims, change claims and their correctness, wrong directions, missed changes and
omissions. Scores are the mean of the two readers per report; differences are paired patient bootstraps (2,000 resamples).
Inter-reader agreement on overall quality: Spearman 0.98, exact 0.92, within one point 1.00.

\begin{table*}[!t]
\centering\small
\caption{\textbf{Reader study} (50 BIND validation pairs, per report). Bold: 95\% CI excludes 0.}
\label{tab:s_readers}
\begin{tabular}{lccc}
\toprule
Per report & Standard format & BrainDiff & Difference [95\% CI] \\
\midrule
Overall quality (1--5) & 2.72 & 3.08 & \textbf{$+$0.36 [$+$0.05, $+$0.68]} \\
Invented findings & 0.68 & 0.39 & $-0.29$ [$-0.80$, $+0.14$] \\
Stable claims wrong & 1.17 & 0.77 & $-0.40$ [$-0.87$, $0.00$] \\
Change claims & 1.32 & 1.43 & $+0.11$ [$-0.99$, $+1.14$] \\
Unsupported change claims & 0.46 & 0.39 & $-0.07$ [$-0.78$, $+0.47$] \\
Correct change claims & 0.23 & 0.47 & $+0.24$ [$-0.08$, $+0.64$] \\
Wrong direction & 0.63 & 0.57 & $-0.06$ [$-0.45$, $+0.35$] \\
Missed changes & 1.24 & 1.15 & $-0.09$ [$-0.33$, $+0.11$] \\
Omitted findings & 0.64 & 0.58 & $-0.06$ [$-0.21$, $+0.08$] \\
Change-claim precision & 0.174 & 0.329 & -- \\
\bottomrule
\end{tabular}
\end{table*}

\paragraph{Reading.}
The status-first model is rated significantly higher overall and makes fewer false stable claims (borderline). Invented
findings and correct, wrong-direction and missed changes do not differ significantly. Readers are LLMs, not radiologists, and the sample is 50 patients.

\section{Fidelity of the Report Rewrites}
\label{suppl:rewrite}

Prior reports (model input) were edited by Claude Opus 5.5 to remove interval-change language. Target reports were edited by
Claude Sonnet 5.5 (BIND: follow-up edit; MR-RATE: restyle) (Sec.~\ref{sec:targets}). We checked every rewrite with Rad-$\Delta$ by comparing the original and the edited report. A status is present when its noisy-OR over the report's findings is at least 0.5. For priors we report the share of reports with any change status (NEW, WORSENED, IMPROVED or RESOLVED) before and after editing. For targets, agreement between original and target on whether any change status is present. \emph{Finding recall}: share
of the original report's abnormal observations (RadGraph-XL, definitely present or uncertain) whose text appears in the edited
report; extracranial findings, which the edit removes by design, count as misses. \emph{Fragment rate}: mean share of Findings
sentences without a finite verb (NLTK part-of-speech proxy; since radiologists also write fragments, compare against the original).

\begin{table*}[!t]
\centering\small
\caption{\textbf{Rewrite fidelity, full sets} (Rad-$\Delta$ labels, original vs.\ edited report).}
\label{tab:s_rewrite}
\begin{tabular}{llcc}
\toprule
 & & BIND & MR-RATE \\
\midrule
\multirow{4}{*}{\shortstack[l]{Priors\\(Opus 5.5)}} & Reports & 27,262 & 14,455 \\
 & Share with any change status, original $\rightarrow$ edit & 0.859 $\rightarrow$ 0.025 & 0.453 $\rightarrow$ 0.050 \\
 & Finding recall & 0.688 & 0.603 \\
 & Fragment rate, original $\rightarrow$ edit & 0.300 $\rightarrow$ 0.144 & 0.042 $\rightarrow$ 0.053 \\
\midrule
\multirow{3}{*}{\shortstack[l]{Targets\\(Sonnet 5.5)}} & Any-change agreement, original vs.\ target & 0.951 & 0.973 \\
 & Finding recall & 0.828 & 0.969 \\
 & Fragment rate, original $\rightarrow$ edit & 0.323 $\rightarrow$ 0.200 & 0.341 $\rightarrow$ 0.417 \\
\bottomrule
\end{tabular}
\end{table*}

\paragraph{Reading.}
The prior edit removes change language from almost all reports (BIND 86\% $\rightarrow$ 2.5\% of reports with a change status).
Its lower finding recall is expected: the most frequent removed findings are change words and normal statements (``unchanged'',
``stable'', ``normal''). Target edits preserve the original report's change status in 95\% (BIND) and 97\% (MR-RATE) of reports,
with per-status agreement 0.964--0.986 (BIND) and 0.964--0.988 (MR-RATE). 

\section{Prompts}
\label{supp:prompts}

All prompts are reproduced verbatim. Text in angle brackets marks inserted content.

\subsection{Generation Prompts}
\label{supp:prompt-generation}

Used to generate the reports of Tab.~\ref{tab:main} (Supp.~\ref{supp:baselines}).

\noindent\textit{BrainDiff.} The first line is the system prompt; each series contributes its visual tokens after its sequence name.
\begin{list}{}{\setlength{\leftmargin}{0.8em}\setlength{\rightmargin}{0pt}\setlength{\topsep}{3pt}}\item\relax\scriptsize\ttfamily\raggedright
You are an expert neuro-radiologist AI assistant. Analyze the provided neuroimaging study and answer the user's request. /no\_think\\
Prior study: T1w \textless{}visual tokens\textgreater{} T1ce \textless{}visual tokens\textgreater{} T2w \textless{}visual tokens\textgreater{} FLAIR \textless{}visual tokens\textgreater{}\\
Current study: T1w \textless{}visual tokens\textgreater{} T1ce \textless{}visual tokens\textgreater{} T2w \textless{}visual tokens\textgreater{} FLAIR \textless{}visual tokens\textgreater{}\\
Prior report: \textless{}edited prior report\textgreater{}\\
The report for the current (follow-up) study is NOT provided. Read the current findings off the images above, and use the prior report only as the baseline to compare against.\\
Write the report for the current study as a follow-up to the prior study, using the following structure: Findings: [current findings, each stated as new, worsened, improved, resolved or unchanged compared with the prior] Impression: [clinical interpretation].
\end{list}

\noindent\textit{Baselines, system prompt (all vision-language baselines).}
\begin{list}{}{\setlength{\leftmargin}{0.8em}\setlength{\rightmargin}{0pt}\setlength{\topsep}{3pt}}\item\relax\scriptsize\ttfamily\raggedright
You are an expert neuroradiologist. You compare a follow-up brain MRI with the prior examination and write the follow-up report.
\end{list}

\noindent\textit{Baselines, instruction for MedGemma-1.5-4B and MedGemma-27B.}
\begin{list}{}{\setlength{\leftmargin}{0.8em}\setlength{\rightmargin}{0pt}\setlength{\topsep}{3pt}}\item\relax\scriptsize\ttfamily\raggedright
Write the FINDINGS section of the report for the CURRENT examination. Describe the intracranial findings on the current scans and, for each relevant finding, state whether it is new, increased/worsened, decreased/improved, resolved, or unchanged compared with the prior examination. Use the prior report as the baseline. Write plain prose without headings or recommendations.
\end{list}

\noindent\textit{Baselines, instruction for Qwen3-VL-8B (released and fine-tuned).}
\begin{list}{}{\setlength{\leftmargin}{0.8em}\setlength{\rightmargin}{0pt}\setlength{\topsep}{3pt}}\item\relax\scriptsize\ttfamily\raggedright
Write the FINDINGS section of the report for the CURRENT examination. Base every comparison on the images: for each finding, look at the PRIOR and CURRENT images of the same sequence and judge from them whether it is new, increased/worsened, decreased/improved, resolved, or unchanged. The prior report tells you what was described on the prior examination; it does not tell you what the current scans show. Do not assume a finding is unchanged, or that it has followed its usual course, unless the current images support it, and report new findings you see on the current images even if the prior report does not mention them. Write plain prose without headings or recommendations.
\end{list}

\subsection{Independent-Labeler System Prompt}
\label{supp:prompt-labeler}

Used by the Qwen3.6-27B status labeler (Supp.~\ref{supp:eval}).

\begin{list}{}{\setlength{\leftmargin}{0.8em}\setlength{\rightmargin}{0pt}\setlength{\topsep}{3pt}}\item\relax\scriptsize\ttfamily\raggedright
You are an expert annotator of radiology report language, annotating interval change in brain MRI radiology reports.\\
\textless{}task\textgreater{} Each case is a PRIOR brain MRI report and a CURRENT brain MRI report of the same patient. Decide which interval-change statuses the CURRENT report states for at least one finding, and output one JSON object: \{"NEW": bool, "WORSENED": bool, "IMPROVED": bool, "RESOLVED": bool\}. A status is true when at least one finding in the CURRENT report carries it under the definitions and rules below; otherwise it is false. Several statuses can be true at once. \textless{}/task\textgreater{}\\
\textless{}core\_principle\textgreater{}\\
LABEL WHAT THE CURRENT REPORT STATES, NOT WHAT IS CLINICALLY TRUE. A status encodes the interval-change language in the CURRENT report text only. Never infer change from clinical knowledge, from the pathology, from the expected natural history of a disease, or from what a prior examination probably showed. THIS IS OF PARAMOUNT IMPORTANCE.\\
The PRIOR report is context only. Use it to understand which findings the CURRENT report refers to (for example, that "the lesion" or "the collection" is a finding already described before). Never assign a status because of wording in the PRIOR report, and never infer change by comparing the two reports yourself: a finding the CURRENT report describes without change language carries no status, even if the PRIOR report did not mention it or described it differently.\\
\textless{}/core\_principle\textgreater{}\\
\textless{}status\_definitions\textgreater{}\\
NEW = the current report states a finding was not present on the prior examination (new, newly seen, interval development, recurrence, now present, has developed).\\
WORSENED = the current report states an existing finding increased in size, number, extent, severity, or conspicuity (increased, enlarged, larger, progression, progressive, worsening, expanded, extending further).\\
IMPROVED = the current report states an existing finding decreased but is still present (decreased, smaller, reduced, regressed, improving, resolving, partial or near-complete resolution).\\
RESOLVED = the current report states a previously seen finding is completely gone (resolved, no longer seen, no longer visualized, complete resolution, interval resolution).\\
Findings stated to be unchanged (stable, unchanged, similar, persistent, again seen, redemonstrated, no significant change, no interval change) and findings without change language carry none of the four statuses.\\
\textless{}/status\_definitions\textgreater{}\\
\textless{}rules\textgreater{}\\
Impression first: when the Findings and the Impression state different changes for the same finding, the Impression applies.\\
Negated change: a negated NEW or WORSENED statement ("no new enhancing lesion", "no progressive mass effect", "without interval enlargement", "no evidence of interval hemorrhage") asserts that nothing appeared or progressed; it makes neither NEW nor WORSENED true. "Has not resolved", "no improvement" -\textgreater{} neither IMPROVED nor RESOLVED.\\
Size-only change: when both a prior and a current measurement are stated for the same finding and no change language is stated for it, derive the status from the numbers: larger -\textgreater{} WORSENED; smaller but present -\textgreater{} IMPROVED; equal -\textgreater{} no status. When explicit change language and stated measurements disagree, use the change language. A current measurement alone never creates a status.\\
Undirected change ("changed in configuration", "evolving", "evolution of", "interval change in appearance") -\textgreater{} no status. Do not guess a direction.\\
Mixed statements: label each finding by the change stated for it ("the left frontal lesion has enlarged while the right parietal lesion has decreased" -\textgreater{} WORSENED and IMPROVED both true).\\
Several cues on one finding: a directional cue beats a persistence cue ("persistent but decreased enhancement" -\textgreater{} IMPROVED; "again seen, slightly larger" -\textgreater{} WORSENED).\\
Fixed terms: a cue word inside a disease name or fixed clinical term (progressive multifocal leukoencephalopathy, progressive supranuclear palsy) is not a change statement.\\
Interval procedures: "interval craniotomy", "interval resection", "interval shunt placement" -\textgreater{} NEW. Postsurgical findings (resection cavity, pneumocephalus, dural enhancement) are NEW only when stated to be new. "Status post", "prior", or "history of" a procedure -\textgreater{} no status. "New postoperative changes" -\textgreater{} NEW; "expected postoperative changes" -\textgreater{} no status.\\
Uncertainty: "possibly new", "questionable increase", "may represent progression" -\textgreater{} NEW / WORSENED / WORSENED. When the report offers alternative changes, use the one that departs from unchanged; if both do, use the first named ("similar to slightly decreased" -\textgreater{} IMPROVED; "new or progressive" -\textgreater{} NEW).\\
Hedged stability ("no definite interval change", "no significant interval change") -\textgreater{} no status, unless a directional alternative is also offered ("no significant change, possibly slightly larger" -\textgreater{} WORSENED).\\
Property, not change: "restricted diffusion", "increased T2 signal", "decreased perfusion", "hyperintense", or "enlarged" used as a morphologic description without comparison ("enlarged ventricles" with no reference to prior) are not change statements. Treat "increased" / "decreased" as change only when the text links them to a prior examination or the sentence is comparative.\\
Several comparison exams: when a finding is compared with more than one prior, use the change relative to the MOST RECENT comparison.\\
Near resolution: RESOLVED only when the finding is stated to be completely gone. "Near-complete resolution", "now barely visible", "now not discretely visualized" -\textgreater{} IMPROVED. "No residual [finding]" -\textgreater{} RESOLVED.\\
Absence: do not infer RESOLVED because a finding is not mentioned; do not infer NEW because a finding is described without a comparison. Comparison-section and history text describe exams, not findings.\\
\textless{}/rules\textgreater{}\\
\textless{}output\_format\textgreater{}\\
Output ONLY the JSON object, with exactly the keys NEW, WORSENED, IMPROVED, RESOLVED in this order and boolean values. No commentary.\\
\textless{}/output\_format\textgreater{}\\
\textless{}examples\textgreater{}\\
PRIOR REPORT: Right temporal enhancing metastasis measuring 1.2 cm. Right frontal subdural collection.\\
CURRENT REPORT: Interval increase in size of the right temporal metastasis, now 1.9 cm, previously 1.2 cm. New 5 mm enhancing focus in the left parietal cortex. Previously seen subdural collection has resolved. Restricted diffusion in the right caudate.\\
Output: \{"NEW": true, "WORSENED": true, "IMPROVED": false, "RESOLVED": true\}\\
\medskip
PRIOR REPORT: Left temporal enhancing lesion. Periventricular white matter lesions.\\
CURRENT REPORT: Left temporal enhancing lesion is similar to slightly decreased in size. No significant change in periventricular white matter lesions, possibly slightly larger.\\
Output: \{"NEW": false, "WORSENED": true, "IMPROVED": true, "RESOLVED": false\}\\
\medskip
PRIOR REPORT: Progressive multifocal leukoencephalopathy. Enhancing right frontal nodule. Small left frontal extra-axial collection.\\
CURRENT REPORT: FINDINGS: Progressive multifocal leukoencephalopathy in the bilateral frontal white matter. No hemorrhage. Previously seen right frontal enhancing nodule is not discretely visualized. No residual extra-axial collection. IMPRESSION: No new hemorrhage.\\
Output: \{"NEW": false, "WORSENED": false, "IMPROVED": true, "RESOLVED": true\}\\
\medskip
PRIOR REPORT: Small chronic infarct in the right cerebellar hemisphere. No mass.\\
CURRENT REPORT: Chronic infarct in the right cerebellar hemisphere. 8 mm enhancing lesion in the left frontal lobe. No new infarct.\\
Output: \{"NEW": false, "WORSENED": false, "IMPROVED": false, "RESOLVED": false\}\\
\textless{}/examples\textgreater{}
\end{list}

\subsection{Reader Brief and Scoring Rubric}
\label{supp:prompt-reader}

Used by the language-model readers (Supp.~\ref{supp:readers}).

\begin{list}{}{\setlength{\leftmargin}{0.8em}\setlength{\rightmargin}{0pt}\setlength{\topsep}{3pt}}\item\relax\scriptsize\ttfamily\raggedright
You are a blinded reader grading machine-generated follow-up brain MRI reports. Each item has:\\
- PRIOR: the report of the prior examination (what the system was given).\\
- REFERENCE: the radiologist's actual report of the CURRENT (follow-up) examination. Treat it as the truth about what changed.\\
- CANDIDATE: a follow-up report written by a system that saw the prior report and the images, but not the REFERENCE.\\
The question is: does the CANDIDATE correctly describe what changed since the prior examination?\\
Only intracranial findings count; ignore extracranial content (bone, orbits, sinuses, mastoids, scalp, neck), technique, history,\\
dates and recommendations, and ignore wording and style.\\
1. Change claims. List every statement in the CANDIDATE that asserts interval change or its absence for an intracranial finding:\\
   new; increased / enlarged / worsened / progressed; decreased / smaller / improved / regressed; resolved / no longer seen;\\
   stable / unchanged / similar / again seen / no new lesion / no interval change. Label each one:\\
   - "correct": the REFERENCE states the same direction for the same finding (or, for stable/no-new claims, does not contradict it).\\
   - "wrong\_direction": the REFERENCE states a different direction for that finding (e.g. candidate says increased, reference\\
     says decreased or stable; candidate says stable, reference says new or worsened).\\
   - "unsupported": the REFERENCE does not mention that finding at all, or gives no basis for the claimed change.\\
2. missed\_changes: the number of interval changes (new / worsened / improved / resolved) stated in the REFERENCE for an\\
   intracranial finding that the CANDIDATE does not state with the correct direction. Stable statements do not count here.\\
3. invented\_findings: the number of intracranial abnormal findings in the CANDIDATE that are in neither the REFERENCE nor the PRIOR.\\
4. omitted\_findings: the number of clinically significant intracranial abnormal findings in the REFERENCE missing from the CANDIDATE.\\
5. change\_score: 1-5, does the CANDIDATE correctly describe what changed since the prior examination, judged against the REFERENCE?\\
   5 = every interval change in the REFERENCE is stated with the correct direction and nothing false is claimed;\\
   4 = the important changes are right, minor omissions or minor false claims;\\
   3 = mixed: some changes right, some important changes missed or wrong;\\
   2 = mostly wrong or missing about change;\\
   1 = misleading about change (e.g. calls progression stable or improvement progression).\\
   If the REFERENCE states no interval change, 5 = the CANDIDATE also states none; lower for each false change claim.\\
Output exactly one JSON object per item, one per line, in item order, with these keys and nothing else:\\
\{"id": "...", "claims": [\{"type": "new|worse|better|resolved|stable", "label": "correct|wrong\_direction|unsupported"\}, ...],\\
 "missed\_changes": 0, "invented\_findings": 0, "omitted\_findings": 0,\\
 "change\_score": 1\}\\
Do not copy report text into the output. Grade each item independently; items are shuffled and come from different systems.
\end{list}

\subsection{Silver-Labeling Prompt}
\label{supp:prompt-silver}

Used to assign the silver labels that train Rad-$\Delta$ (Supp.~\ref{suppl:silver}).

\paragraph{System prompt (brain MRI).}
\begin{list}{}{\setlength{\leftmargin}{0.8em}\setlength{\rightmargin}{0pt}\setlength{\topsep}{3pt}}\item\relax\scriptsize\ttfamily\raggedright
You are an expert annotator of radiology report language, annotating interval change in brain MRI radiology reports for machine learning training data.\\
\textless{}task\textgreater{} Each case is ONE brain MRI radiology report in which every RadGraph Observation entity has been marked inline with a numeric id, e.g. "unchanged[3] T2 prolongation[4]". For every marked Observation entity, assign exactly ONE change attribute describing what the report states about that entity's interval change relative to any prior examination. Do not rewrite the report. Do not add, remove, merge, or split entities. Do not label unmarked text. The entity boundaries and ids are fixed; your only output is one attribute per id. \textless{}/task\textgreater{}\\
\textless{}core\_principle\textgreater{}\\
LABEL WHAT THE REPORT STATES, NOT WHAT IS CLINICALLY TRUE. The attribute encodes the interval-change language in the report text only. Never infer change from clinical knowledge, from the pathology, from the expected natural history of a disease, or from what a prior examination probably showed. If the report does not express interval change for an entity, the attribute is NONE. THIS IS OF PARAMOUNT IMPORTANCE.\\
\textless{}/core\_principle\textgreater{}\\
\textless{}attribute\_definitions\textgreater{}\\
The attribute must be exactly one of: NEW, WORSENED, IMPROVED, STABLE, RESOLVED, NONE\\
NEW = the report states the finding was not present on the prior examination (new, newly seen, interval development, recurrence, now present, has developed).\\
WORSENED = the report states an existing finding increased in size, number, extent, severity, or conspicuity (increased, enlarged, larger, progression, progressive, worsening, expanded, extending further).\\
IMPROVED = the report states an existing finding decreased but is still present (decreased, smaller, reduced, regressed, improving, resolving, partial or near-complete resolution).\\
STABLE = the report states the finding is unchanged relative to prior (stable, unchanged, similar, persistent, again seen, redemonstrated, no significant change, no interval change).\\
RESOLVED = the report states a previously seen finding is completely gone (resolved, no longer seen, no longer visualized, complete resolution, interval resolution).\\
NONE = the report expresses no interval change for this entity.\\
\textless{}/attribute\_definitions\textgreater{}\\
\textless{}attribute\_rules\textgreater{}\\
Precedence: (1) change stated in the Impression about this finding; (2) change stated about this entity; (3) change stated about the same finding elsewhere in the report (mentions rule); (4) NONE only when none of these applies. Exceptions: a "status post" / "prior" / "history of" procedure mention is NONE (see Interval procedures); comparison-section and history text is NONE (see absence and validation rules); a negated NEW or WORSENED statement never propagates to another mention (see Negated change).\\
Head and cue entities: RadGraph often marks the change word itself as its own Observation entity (e.g. "unchanged[3]", "new[7]", "increase[9]") and links it to the finding it modifies. Label BOTH the change-word entity and the finding it describes with the same attribute. Example: "unchanged[3] T2 prolongation[4]" -\textgreater{} 3=STABLE; 4=STABLE.\\
Descriptors inherit: every entity that describes a finding -- severity (mild, moderate), size or count words (small, punctate, multiple, numerous), morphology or signal (rim enhancing, calcified, nodular), chronicity (chronic, acute), properties (size, extent), and measurements (6 mm) -- receives the same attribute as the finding it describes. Treat the change word, the finding, and every entity describing that finding as one group; all entities in the group receive the same attribute. Examples: "Numerous[1] rim enhancing[2] lesions[3] have increased[4] in size[5]" -\textgreater{} 1=WORSENED; 2=WORSENED; 3=WORSENED; 4=WORSENED; 5=WORSENED. "unchanged[1] partially[2] calcified[3] nodule[4]" -\textgreater{} 1=STABLE; 2=STABLE; 3=STABLE; 4=STABLE. A descriptor that governs findings with different attributes -\textgreater{} NONE.\\
Fixed terms, not change: a cue word creates a change attribute only when it asserts change relative to a prior examination. When it is part of a disease name or fixed clinical term (progressive multifocal leukoencephalopathy, progressive supranuclear palsy), it is a descriptor and inherits the attribute of the finding it describes.\\
Mentions of the same finding: RadGraph marks each mention separately, so one finding (same pathology, same location) may carry several ids, e.g. in Findings and Impression. When the report states change for a finding in any mention, assign that attribute to every mention of that finding, including mentions with no local change language. If mentions state different changes, the change stated in the Impression applies to every mention; if the Impression states no change for the finding, each mention keeps its own stated change and mentions without change language take the Findings statement.\\
Negated change: a negated NEW or WORSENED statement asserts that the finding did not appear or did not progress; label it STABLE. Examples: "no new[5] enhancing lesion[6]" -\textgreater{} 5=STABLE; 6=STABLE. "no progressive[2] mass effect[3]" -\textgreater{} 2=STABLE; 3=STABLE. "without interval enlargement[4]" -\textgreater{} 4=STABLE. "Interval" means new since the prior, so a negated "interval" is a negated NEW: "no evidence of interval[1] intracranial hemorrhage[2] or acute infarction[3]" -\textgreater{} 1=STABLE; 2=STABLE; 3=STABLE. A negated NEW or WORSENED statement never assigns an attribute to any other mention, including through the Impression or the mentions rule; a mention without its own change language stays NONE. Example: "No hemorrhage[1]. IMPRESSION: No new[2] hemorrhage[3]." -\textgreater{} 1=NONE; 2=STABLE; 3=STABLE. Presence and absence are already encoded by RadGraph and must not influence this rule.\\
Negated resolution or improvement: "has not resolved", "no improvement" -\textgreater{} STABLE.\\
Size-only change: when both a prior and a current measurement are stated for the same finding and no change language is stated for it, derive the attribute from the numbers: larger -\textgreater{} WORSENED; smaller but present -\textgreater{} IMPROVED; equal -\textgreater{} STABLE. When explicit change language and stated measurements disagree, use the change language. A current measurement alone never creates a change attribute; the measurement entity inherits the attribute of the finding it describes, which is NONE if nothing else states a change.\\
Similar: "similar", "similar in size", "similar appearance" -\textgreater{} STABLE.\\
Undirected change: when the report states change without a direction ("changed in configuration", "evolving", "evolution of", "interval change in appearance") -\textgreater{} NONE. Do not guess a direction.\\
Mixed statements: label each entity by the change stated for that entity. "The left frontal lesion has enlarged while the right parietal lesion has decreased" -\textgreater{} left frontal lesion WORSENED; right parietal lesion IMPROVED.\\
Several cues on one entity: when one entity has both a persistence cue (persistent, again seen, redemonstrated, still) and a directional cue, use the directional cue. "persistent but decreased enhancement" -\textgreater{} IMPROVED; "again seen, slightly larger" -\textgreater{} WORSENED.\\
Shared change words: when one change word governs several findings ("unchanged ventricles and sulci", "no new infarct or hemorrhage"), every governed finding receives the attribute. A negated change word distributes across the whole list it governs: "no evidence of a new[1] intracranial mass[2], intracranial hemorrhage[3] or acute infarction[4]" -\textgreater{} 1=STABLE; 2=STABLE; 3=STABLE; 4=STABLE.\\
Interval procedures: "interval [procedure]" ("interval craniotomy", "interval resection", "interval shunt placement") -\textgreater{} NEW for the procedure entity and its descriptors only. Postsurgical findings (resection cavity, pneumocephalus, dural enhancement) are NEW only when the report states they are new; otherwise they follow the other rules. "Status post", "prior", or "history of" a procedure without "interval" -\textgreater{} NONE, even when another mention of the same procedure says "interval"; the mentions rule does not apply to these mentions.\\
Uncertainty is orthogonal: "possibly new", "questionable increase", "may represent progression" -\textgreater{} NEW / WORSENED / WORSENED. RadGraph encodes uncertainty separately. When the report offers alternative changes, use the one that departs from STABLE; if both do, use the first named: "similar to slightly decreased" -\textgreater{} IMPROVED; "new or progressive" -\textgreater{} NEW.\\
Property, not change: words that describe the tissue or signal rather than a change over time are not change cues by themselves: they never create a change attribute. When they describe a finding that carries a change attribute, they inherit it like any other descriptor ("new[1] restricted diffusion[2]" -\textgreater{} 1=NEW; 2=NEW). Otherwise they follow the precedence order, including the mentions rule; they are NONE only when no mention of the finding states change. "decreased diffusivity", "restricted diffusion", "increased T2 signal", "decreased perfusion", "hyperintense", "enlarged" used as a morphologic description without comparison ("enlarged ventricles" with no reference to prior) do not create change on their own. Treat "increased"/"decreased" as change only when the text links them to a prior examination or the surrounding sentence is comparative.\\
Several comparison exams: when the report compares one finding against more than one prior ("new since the prior MRI but unchanged since the recent CT"), use the change relative to the MOST RECENT comparison -\textgreater{} STABLE in that example.\\
Hedged stability: "without convincing evidence of interval change", "no definite interval change", "no significant interval change" -\textgreater{} STABLE. Hedging does not make it NONE. When a hedged stability statement also offers a directional alternative, the Uncertainty rule applies: "no significant change, possibly slightly larger" -\textgreater{} WORSENED.\\
Near resolution: RESOLVED only when the finding is stated to be completely gone. "Near-complete resolution", "now barely visible", "now not discretely visualized" said of a previously seen finding -\textgreater{} IMPROVED, whether or not a decrease is stated. "No residual [finding]" -\textgreater{} RESOLVED.\\
Postoperative and treatment findings follow the same rules: "new postoperative changes" -\textgreater{} NEW; "expected postoperative changes" -\textgreater{} NONE.\\
\textless{}/attribute\_rules\textgreater{}\\
\textless{}output\_format\textgreater{}\\
Produce ONLY lines matching this exact format. Make sure every report number matches the input case number:\\
Report \{n\}: 1=ATTRIBUTE; 2=ATTRIBUTE; ...; k=ATTRIBUTE\\
Every marked id in the case must appear exactly once, in ascending order. No blank lines. No line breaks within a report. No commentary. No explanations. Each report must be a single continuous line.\\
Examples:\\
Input:\\
Case 12\\
Report: Unchanged[1] mild[2] T2 prolongation[3] in the left frontal white matter. No new[4] enhancing lesion[5]. Ventricles are normal[6] in size.\\
Output: Report 12: 1=STABLE; 2=STABLE; 3=STABLE; 4=STABLE; 5=STABLE; 6=NONE\\
Input:\\
Case 13\\
Report: Interval increase[1] in size[2] of the right temporal metastasis[3], now 1.9 cm, previously 1.2 cm. New[4] 5 mm enhancing focus[5] in the left parietal cortex. Previously seen[6] subdural collection[7] has resolved[8]. Restricted diffusion[9] in the right caudate.\\
Output: Report 13: 1=WORSENED; 2=WORSENED; 3=WORSENED; 4=NEW; 5=NEW; 6=RESOLVED; 7=RESOLVED; 8=RESOLVED; 9=NONE\\
\textless{}/output\_format\textgreater{}\\
\textless{}examples\textgreater{}\\
Impression wins over a conflicting Findings statement:\\
Input:\\
Case 14\\
Report: FINDINGS: Slightly increased[1] FLAIR hyperintensity[2] in the right frontal lobe. IMPRESSION: Right frontal FLAIR hyperintensity[3], unchanged[4].\\
Output: Report 14: 1=STABLE; 2=STABLE; 3=STABLE; 4=STABLE\\
Interval procedure labels the procedure only:\\
Input:\\
Case 15\\
Report: Interval right frontal craniotomy[1] and resection[2]. Resection cavity[3] with small[4] pneumocephalus[5].\\
Output: Report 15: 1=NEW; 2=NEW; 3=NONE; 4=NONE; 5=NONE\\
Alternatives, including hedged stability with a directional alternative:\\
Input:\\
Case 16\\
Report: Left temporal enhancing lesion[1] is similar[2] to slightly decreased[3] in size[4]. No significant change[5] in periventricular white matter lesions[6], possibly slightly larger[7].\\
Output: Report 16: 1=IMPROVED; 2=IMPROVED; 3=IMPROVED; 4=IMPROVED; 5=WORSENED; 6=WORSENED; 7=WORSENED\\
Cue word inside a disease name; directional cue beats persistence cue:\\
Input:\\
Case 18\\
Report: Progressive[1] multifocal leukoencephalopathy[2] in the bilateral frontal white matter. Persistent[3] but decreased[4] enhancement[5] in the right parietal lobe. Right occipital lesion[6] is again seen[7] and slightly larger[8].\\
Output: Report 18: 1=NONE; 2=NONE; 3=IMPROVED; 4=IMPROVED; 5=IMPROVED; 6=WORSENED; 7=WORSENED; 8=WORSENED\\
Negated change does not propagate; not discretely visualized; no residual:\\
Input:\\
Case 19\\
Report: FINDINGS: No hemorrhage[1]. Previously seen[2] right frontal enhancing nodule[3] is not discretely visualized[4]. No residual[5] extra-axial collection[6]. IMPRESSION: No new[7] hemorrhage[8].\\
Output: Report 19: 1=NONE; 2=IMPROVED; 3=IMPROVED; 4=IMPROVED; 5=RESOLVED; 6=RESOLVED; 7=STABLE; 8=STABLE\\
Change word beats measurements; property words inherit through mentions:\\
Input:\\
Case 20\\
Report: FINDINGS: Stable[1] 1.2 cm[2] left parietal meningioma[3], previously 1.1 cm[4]. Increased[5] T2 signal[6] in the left centrum semiovale. IMPRESSION: Unchanged[7] left centrum semiovale T2 signal abnormality[8].\\
Output: Report 20: 1=STABLE; 2=STABLE; 3=STABLE; 4=STABLE; 5=STABLE; 6=STABLE; 7=STABLE; 8=STABLE\\
\textless{}/examples\textgreater{}\\
\textless{}absence\_and\_validation\_rules\textgreater{}\\
Do not infer RESOLVED because a finding is simply not mentioned; RESOLVED requires explicit language.\\
Do not infer NEW because a finding is described without a comparison; NEW requires explicit language ("new", "newly", "interval development", "not previously seen").\\
A finding described as "again seen" or "redemonstrated" without any size or severity change is STABLE.\\
Comparison-section text ("Comparison: MRI 2019") and history text describe exams, not findings; entities there are NONE unless they state change in a finding.\\
\textless{}/absence\_and\_validation\_rules\textgreater{}\\
\textless{}error\_handling\textgreater{}\\
If mentions of the same finding state different changes, the change stated in the Impression applies to every mention (see Mentions of the same finding).\\
If a report contains no change language at all, every id is NONE.\\
\textless{}/error\_handling\textgreater{}\\
\textless{}validation\_checklist\textgreater{}\\
Before outputting each report, verify internally:\\
FORMAT: Output matches exactly "Report \{n\}: 1=ATTRIBUTE; 2=ATTRIBUTE; ..."; every marked id appears exactly once in ascending order; every attribute is exactly one of NEW, WORSENED, IMPROVED, STABLE, RESOLVED, NONE; no blank lines; no commentary.\\
ACCURACY: Every non-NONE attribute is supported by explicit change language in the report; negated NEW/WORSENED is STABLE and does not propagate to other mentions; tissue or signal properties and cue words inside disease names never create change on their own; cue entities carry the same attribute as the finding they modify; descriptors and measurements carry the attribute of the finding they describe; the Impression's stated change applies to every mention of that finding; directional cues beat persistence cues; change language beats measurements; nothing is inferred from clinical knowledge.\\
\textless{}/validation\_checklist\textgreater{}\\
\textless{}input\_format\textgreater{}\\
Input may contain one or more cases. Each case contains a case number and one report with inline entity ids. Example:\\
Case 1\\
Report: ...\\
Case 2\\
Report: ...\\
The output must include one line per case in the same order. Keep \{n\} aligned with the input case number.\\
\textless{}/input\_format\textgreater{}\\
\end{list}
\paragraph{User turn.} Eight cases per request; angle brackets mark inserted content.
\begin{list}{}{\setlength{\leftmargin}{0.8em}\setlength{\rightmargin}{0pt}\setlength{\topsep}{3pt}}\item\relax\scriptsize\ttfamily\raggedright
Process the cases below.\\
Each case is one brain MRI report with inline Observation entity ids.\\
\#\#\# CASES (BEGIN NOW)\\
Case \textless{}n\textgreater{}\\
Report: \textless{}report with Observation entities marked [1]...[k]\textgreater{}\\
\end{list}
\paragraph{Chest X-ray.} The chest X-ray prompt is a copy of the brain prompt with the modality name and the example findings changed to chest findings. Precedence, attribute set, shared rules and output format are identical. It adds these rules at the end of the attribute rules:
\begin{list}{}{\setlength{\leftmargin}{0.8em}\setlength{\rightmargin}{0pt}\setlength{\topsep}{3pt}}\item\relax\scriptsize\ttfamily\raggedright
Support devices (endotracheal and enteric tubes, central lines, PICC lines, chest tubes, pacemaker and defibrillator leads, sternal wires, valves, stents): "interval placement of", "new" device -\textgreater{} NEW; "removed", "interval removal", "no longer seen" device -\textgreater{} RESOLVED; "unchanged position", "stable support devices", "remains in place" -\textgreater{} STABLE. "in place" alone, with no comparison, is NONE. Repositioning without a better/worse statement ("advanced", "retracted", "now terminates in", "repositioned") -\textgreater{} NONE (undirected change).\\
Graded findings: a change in stated grade is directional, like a size change. "small, previously moderate" or "now trace" from small -\textgreater{} IMPROVED; "moderate, previously small" -\textgreater{} WORSENED; same grade stated for both -\textgreater{} STABLE.\\
Aeration and clearing: "improved aeration" of a region means the opacity or atelectasis there IMPROVED; "worsened aeration" -\textgreater{} WORSENED; "clearing" -\textgreater{} IMPROVED; "cleared" -\textgreater{} RESOLVED.\\
Persistence wording: "remains", "continues to", "persists", "still" -\textgreater{} STABLE, unless a directional cue is also stated (Several cues on one entity).\\
Technique, not change: a difference attributed to technique, projection, positioning, or inspiratory effort ("lower lung volumes likely due to technique", "differences in positioning", "portable technique limits comparison") -\textgreater{} NONE.\\
\end{list}

\end{document}